\documentclass[11pt]{article}

\usepackage[preprint]{acl}

\usepackage{times}
\usepackage{latexsym}

\usepackage[T1]{fontenc}
\usepackage[utf8]{inputenc}

\usepackage{microtype}

\usepackage{inconsolata}

\usepackage{graphicx}
\graphicspath{{paper_figures/}}  
\usepackage{placeins}            
\usepackage{float}               


\usepackage{amsmath}
\usepackage{amssymb}

\usepackage{booktabs}
\usepackage{multirow}

\newcommand{\rdn}[1]{{\scriptsize\textcolor{red}{(#1$\downarrow$)}}}
\newcommand{\rup}[1]{{\scriptsize\textcolor{green!55!black}{(#1$\uparrow$)}}}

\usepackage{tcolorbox}
\tcbuselibrary{skins,breakable}

\usepackage{listings}
\lstdefinestyle{prompt}{
  basicstyle=\ttfamily\small,
  breaklines=true,
  breakatwhitespace=true,
  columns=fullflexible,
  keepspaces=true,
  showstringspaces=false,
  frame=single,
  framerule=0.4pt,
  framesep=4pt,
  xleftmargin=0pt,
  xrightmargin=0pt,
  aboveskip=6pt,
  belowskip=6pt,
  literate={`}{\textasciigrave}1
}

\title{CAVEWOMAN: How Large Language Models Behave Under Linguistic Input and Output Compression}

\author{Morayo Danielle Adeyemi \\
  Independent \\
  \texttt{morayo.danielle@gmail.com} \\\And
  Ryan A.\ Rossi \\
  Adobe Research \\
  \texttt{ryrossi@adobe.com} \\\And
  Franck Dernoncourt \\
  Adobe Research \\
  \texttt{franck.dernoncourt@adobe.com} \\}

\begin{document}
\maketitle

\begin{abstract}
\textit{``Talk short. Drop grammar. Save token.''} This caveman style is widely promoted as a way to cut inference cost, but whether it actually saves anything depends on which channel (the user's prompt or the model's response) is being compressed. We present \textsc{Cavewoman}, a two-channel evaluation protocol that scores every generation on task accuracy, realized per-item cost, and reference-text agreement against the model's unconstrained reference. We evaluate eight models on five datasets at five reduction levels, with both channels measured on the same items. Output compression cuts realized cost on most API models (\textbf{1.4--2.4$\times$} per model, up to \textbf{3$\times$} in the best case) and on all four open-weight models under public-tier pricing. Input compression has the opposite effect, a strict lose-lose: it raises net cost rather than lowering it (\textbf{$\approx 1.15\times$} on the five-benchmark mean, up to \textbf{1.8$\times$} on the worst dataset and \textbf{2.7$\times$} under stronger compression), because models compensate with longer responses even as accuracy collapses. Under the same setting, surface text diverges from the unconstrained reference: on the non-reasoning models, roughly half of all generations are correct yet their surface text no longer entails the model's own unconstrained baseline generation. The divergence survives length-controlled re-scoring, multiple-comparisons correction, and replication under complementary semantic measures. Code and data are available at \url{https://github.com/danielle34/cavewoman}.
\end{abstract}

\begin{figure*}[t]
\centering
\includegraphics[width=\linewidth]{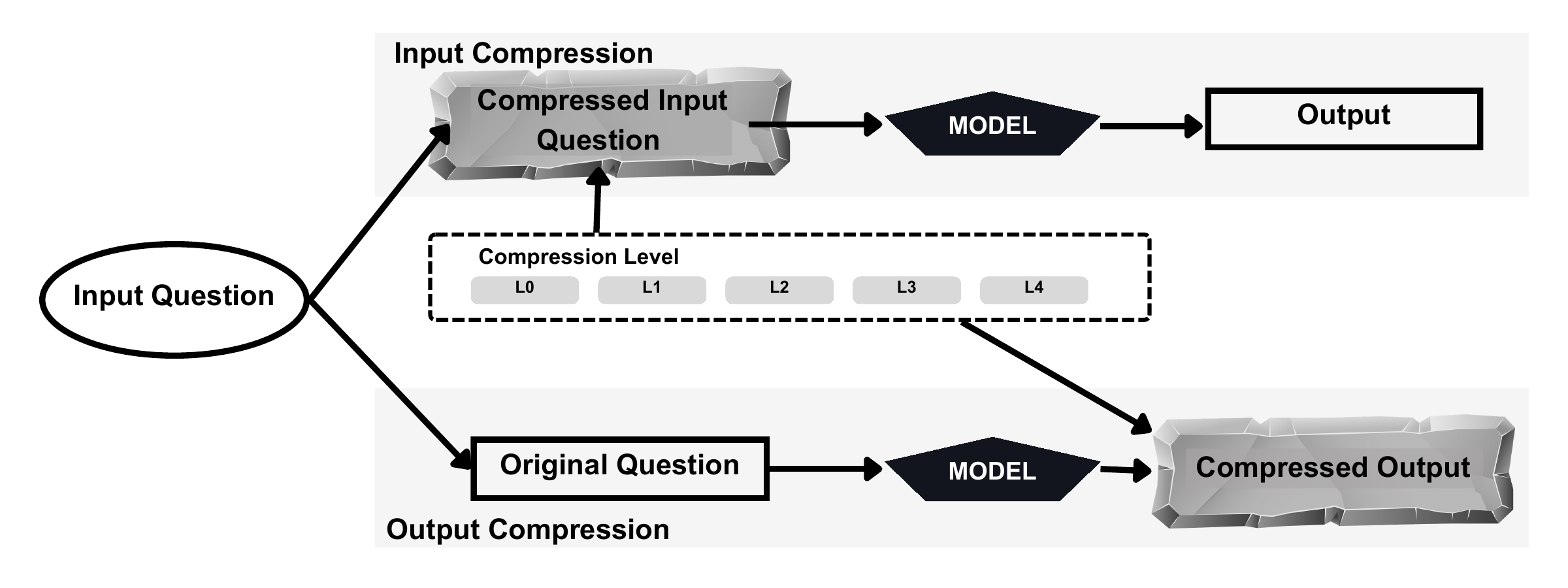}
\caption{\textsc{Cavewoman} framework. The \textbf{input-compression} channel applies a deterministic part-of-speech filter to the user prompt at five reduction levels, leaving the system prompt fixed. The \textbf{output-compression} channel leaves the prompt verbatim and replaces the system prompt with a level-specific instruction that requires the same reduction in the response. Every generation is scored on task accuracy, reference-text agreement against the model's unconstrained reference under bidirectional NLI (plus eleven complementary measures), and per-item input/output token cost.}
\label{fig:framework}
\end{figure*}

\section{Introduction}
\label{sec:intro}

Inference cost in large language models scales with both input and output token counts, and output tokens are typically priced 4--8$\times$ higher than input tokens \citep{ahia2023do, nag-etal-2024-cost}. Existing compression methods reduce one side or the other, either the prompt \citep{jiang-etal-2023-llmlingua, pan2024llmlingua, brussee2026caveman, pelser2026cavemancompression} or the response \citep{xia2025tokenskip, song2025hansel}, but the two are studied separately and both are evaluated almost entirely through task accuracy at a reduced token count. Accuracy at a token count is the wrong instrument for two reasons. It does not separate realized cost from prompt-token reduction (a shorter prompt does not reduce realized cost when the model answers at greater length), and it collapses each generation to a binary outcome that cannot distinguish a compressed answer that agrees with the model's unconstrained reasoning from one that diverges from it.

We address both gaps with \textsc{Cavewoman}, a two-channel evaluation protocol that scores every generation on three axes: task accuracy, realized per-item cost on the priced channel, and reference-text agreement against the model's own unconstrained generation. The protocol measures \emph{input compression} (the prompt is filtered before the model sees it) and \emph{output compression} (the model is instructed to answer in a constrained register) at five reduction levels, holding model and item fixed. We evaluate eight models (Qwen2.5-VL-7B, Qwen3.5-9B, DeepSeek-R1-Distill-Qwen-7B, Gemma-4-E4B, GPT-4o, GPT-5.4, Claude Haiku~4.5, Claude Sonnet~4.6) on five benchmarks (GSM8K, BoolQ, ARC-Easy, CommonsenseQA, MMLU-STEM) with complete coverage of both channels at all five levels.

\paragraph{Contributions.}
\begin{enumerate}
\setlength{\itemsep}{2pt}
\item We propose a two-channel evaluation protocol that scores compression on realized cost with audited answer-extraction rates and a twelve-metric semantic battery (\S\ref{sec:method}).
\item We measure cost asymmetry between input and output compression on the same items (\S\ref{sec:finding-cost}).
\item We measure a surface-text divergence between correct answers and the model's unconstrained reference under output compression, replicated across complementary semantic measures (\S\ref{sec:finding-dissociation}).
\end{enumerate}

\section{Related Work}
\label{sec:related-work}

\paragraph{Input compression.}
Input compression splits along a hard--soft axis \citep{li2025prompt}. Hard methods prune tokens by self-information \citep{li2023compressing}, perplexity \citep{jiang-etal-2023-llmlingua}, question-aware scoring \citep{jiang2024longllmlingua}, or distilled token classification \citep{pan2024llmlingua}, with the same logic at the document level via retrieved-context summarization \citep{xu2024recomp}. Soft methods encode prompts as gist tokens \citep{mu2023learning}, recursive summary vectors \citep{chevalier2023adapting}, or autoencoder slots \citep{ge2024incontext}, with rate--distortion bounds in \citet{nagle2024fundamental}. None measures whether the response says the same thing it would have without compression.

\paragraph{Output compression.}
Output-side methods constrain generation via modified decoding (length controls \citep{kikuchi2016controlling}, budget-signalling positional encodings \citep{takase-okazaki-2019-positional}, countdown mechanisms \citep{song2025hansel}) or via post-training and prompting (token-skipping chains \citep{xia2025tokenskip}, cognitive-inspired routing \citep{aytes2025sketch}, difficulty-aware prompting \citep{han2025token}, RL-driven demonstration compression \citep{huang2024fewer}). The content/function-word split the caveman register exploits is long-standing in linguistics, but this line of work still reports only task accuracy at a token budget.

\paragraph{Semantic fidelity, verbosity, and cost.}
Bidirectional NLI entailment separates propositional content from lexical form \citep{kuhn2023semantic} and has scored summarization faithfulness \citep{maynez2020faithfulness} and inter-system consistency \citep{laban2022summac}; relatedly, chains of thought can be plausible yet causally disconnected from the prediction \citep{turpin2023language, lanham2023measuring}. Length itself matters: verbose outputs score lower \citep{zhang2025demystify}, verbosity carries cost \citep{borisov2026chatbot}, irrelevant padding degrades reasoning \citep{levy2024same}, and chain-of-thought length tracks accuracy independently of trace correctness \citep{jin2024impact, sun2025empirical, wang2024reasoning}. The cost--quality frontier is itself a design surface, in selective routing \citep{chen2024frugalgpt}, quality thresholds \citep{ding2024hybrid}, and preference routers \citep{ong2025routellm}; valid benchmarking calls for multi-metric measurement \citep{bowman2021will, liang2023holistic, gehrmann2021gem}.

\paragraph{Positioning.}
The closest prior work compresses one side only: LLMLingua \citep{jiang-etal-2023-llmlingua, pan2024llmlingua, jiang2024longllmlingua} on the input; TokenSkip \citep{xia2025tokenskip}, Hansel \citep{song2025hansel}, and Sketch-of-Thought \citep{aytes2025sketch} on the output, all reporting task accuracy at compressed token budgets. \textsc{Cavewoman} measures both channels on the same items, reports realized per-item cost rather than token reduction, audits answer-extraction rates before any accuracy claim, and scores reference-text agreement under complementary semantic criteria. The input-channel divergence reproduces under LLMLingua-2 (Appendix~\ref{app:llmlingua}).

\section{Methodology}
\label{sec:method}

\subsection{Experimental Design}
\label{sec:conditions}

\paragraph{Setup.}
Let $\mathcal{M}$ be a fixed language model and let $x=(w_1,\dots,w_n)$ be a
question, a sequence of $n$ tokens from the $\texttt{spaCy}$ tokenizer. Each
token $w$ carries a Penn Treebank part-of-speech tag $g(w)$ assigned by
$\texttt{spaCy}$ with the \texttt{en\_core\_web\_sm} model; $g$ is
deterministic given a fixed $\texttt{spaCy}$ version and model.

For a set $S$ of part-of-speech tags, write $\big(w_i\big)_{\,i\,:\,g(w_i)\in S}$
for the subsequence of $x$ that retains exactly the tokens whose tag lies in
$S$, with the indices $i$ taken in increasing order. Let $\mathrm{trunc}_k(z)$
denote the prefix of a sequence $z$ of length $\min(|z|,k)$, and let
$\texttt{NN}{*}$ and $\texttt{VB}{*}$ denote the Penn Treebank noun and verb tag
families.

We study a single reduction parameter, the level $\ell\in\{0,1,2,3,4\}$, named
L0 through L4. The level selects one filter from the family
$\phi_0,\dots,\phi_4$ defined by
\begin{equation}
\begin{aligned}
\phi_0(x) &= x, \\[2pt]
\phi_1(x) &= \big(w_i\big)_{\,i\,:\,g(w_i)\notin\{\texttt{DT},\texttt{IN},
             \texttt{CC},\texttt{RP},\texttt{TO},\texttt{MD}\}}, \\[2pt]
\phi_2(x) &= \big(w_i\big)_{\,i\,:\,g(w_i)\in\,\texttt{NN}{*}\,\cup\,
             \texttt{VB}{*}\,\cup\,\{\texttt{CD}\}}, \\[2pt]
\phi_3(x) &= \big(w_i\big)_{\,i\,:\,g(w_i)\in\,\texttt{NN}{*}\,\cup\,
             \{\texttt{CD}\}}, \\[2pt]
\phi_4(x) &= \mathrm{trunc}_{15}\big(\phi_3(x)\big).
\end{aligned}
\end{equation}
The family is nested: for every question $x$ and every $\ell\ge 1$,
$\phi_\ell(x)$ is a subsequence of $\phi_{\ell-1}(x)$; a larger $\ell$
thus applies a stricter reduction. The box below gives the linguistic
interpretation of each level.

\paragraph{Two conditions.}
\textsc{Cavewoman} holds $\mathcal{M}$ and $x$ fixed and applies the same
reduction $\phi_\ell$ at one of two points (Figure~\ref{fig:framework}). The
question and the system prompt are formatted with the model's chat template,
held fixed throughout. The L0 baseline is condition-specific: Condition~A uses
the neutral system prompt below, while Condition~B uses the unconstrained
step-by-step prompt of Appendix~\ref{app:condB}; this difference is by design
and later matters for the L0-A/L0-B noise-floor limitation.

\textbf{Condition A (input compression).} The model receives the filtered
question $\phi_\ell(x)$ under a neutral system prompt. Condition~A tests whether
$\mathcal{M}$ needs a full grammatical question.

\textbf{Condition B (output constraint).} The model receives the unmodified
question $x$, and the system prompt instructs it to answer in the reduced form
that $\phi_\ell$ produces. Prompts are task-neutral: $\mathcal{M}$ infers the
answer format from $x$ and a final-line \texttt{Answer:~<answer>} convention.
Condition~B tests whether $\mathcal{M}$ needs expressive freedom in its
response.

The level $\ell$ thus indexes both conditions through one family of filters:
Condition~A reduces what the model reads, and Condition~B reduces what the model
may write.

\subsection{Levels of Linguistic Reduction}
\label{sec:levels}

We define five linguistic reduction levels, with short names we use throughout the paper: the \emph{unconstrained baseline} (L0), the \emph{telegraphic} register (L1), the \emph{keyword-only} register (L2), the \emph{noun-phrase skeleton} (L3), and the \emph{15-token budget} (L4). Each level removes further word classes from the previous level, forming a monotone hierarchy in which a change in accuracy between adjacent levels is attributable to the words removed at that step. The POS filter is a deterministic, transparent reduction chosen for measurement clarity; learned compressors with task-aware token scoring (Appendix~\ref{app:llmlingua}) make different trade-offs.

\begin{tcolorbox}[
  colback=gray!8, colframe=gray!45, boxrule=0.5pt, arc=2pt,
  title={\textbf{Reduction Levels}},
  coltitle=black, fonttitle=\bfseries,
  left=6pt, right=6pt, top=4pt, bottom=4pt,
  breakable]
\begin{description}
\setlength{\itemsep}{2pt}
\item[\textbf{L0: Unconstrained baseline.}]
$\phi_0$ is the identity and keeps every tag. Condition~A presents $x$ verbatim,
and Condition~B asks for step-by-step reasoning in full sentences.
\item[\textbf{L1: Telegraphic.}]
$\phi_1$ removes the closed-class tags \texttt{DT}, \texttt{IN}, \texttt{CC},
\texttt{RP}, \texttt{TO}, and \texttt{MD}: determiners, prepositions,
conjunctions, particles, infinitival \emph{to}, and modals.
\item[\textbf{L2: Keyword only.}]
$\phi_2$ keeps only nouns, verbs, and cardinal numbers.
\item[\textbf{L3: Noun-phrase skeleton.}]
$\phi_3$ keeps only nouns and cardinal numbers.
\item[\textbf{L4: 15-token budget.}]
$\phi_4$ applies $\phi_3$ and then truncates the result to its first $15$
tokens.
\end{description}
\end{tcolorbox}

The same five levels apply to both conditions but enter the pipeline at different points. In Condition~A (input compression), $\phi_\ell$ rewrites the user message via the deterministic POS-tag filter. In Condition~B (output constraint), a level-specific system prompt instructs the model to produce its response in the matching register; the user message is left intact. The per-level decoder budget is $\texttt{max\_new\_tokens}~\in\{400, 300, 200, 150, 20\}$ across L0--L4, identical for both conditions. Both the verbatim per-level filter rules (Condition~A) and the verbatim per-level system prompts (Condition~B) are listed in Appendix~\ref{app:condA}. L4 is excluded from reference-text-agreement scoring since its Condition~B system prompt asks the model to emit only the answer; the 15-token budget is conveyed as a prompt instruction rather than a hard decoder cap, and its soft-enforcement details are in Appendix~\ref{app:extraction_audit}. A worked example of one question filtered through all five Condition~A levels is in Appendix~\ref{app:condA} (Figure~\ref{fig:input_compression}).

\subsection{Evaluation Metrics}
\label{sec:metrics}

We score every generation on three axes: task accuracy (the regex-extracted answer matches the item's ground-truth answer, with a 0.01 tolerance on numeric answers), per-item realized token cost on the priced channel, and reference-text agreement against the model's unconstrained generation on the same item and channel. Reference-text agreement is operationalized by bidirectional NLI entailment, with a DeBERTa-based NLI judge \citep{he2020deberta} as the conservative headline criterion, and replicated under eleven complementary semantic criteria (Appendix~\ref{app:semantic_robustness}).

\subsection{Dissociation Table}
\label{sec:dissociation}

For each level $L_x$ we build a $2 \times 2$ table that crosses task correctness with \emph{reference-text agreement against L0} (Table~\ref{tab:dissociation}). $C_2$ and $C_3$ are distinct outcomes that accuracy-only evaluation cannot separate; outcome shares are robust to metric choice (Appendix~\ref{app:semantic_robustness}).

\begin{table}[!htbp]
\centering
\small
\begin{tabular}{lcc}
\toprule
 & \textbf{Entails L0} & \textbf{Does not entail L0} \\
\midrule
\textbf{Correct}   & $C_1$ & $C_2$ \\
\textbf{Incorrect} & $C_3$ & $C_4$ \\
\bottomrule
\end{tabular}
\caption{$2 \times 2$ dissociation table, applied at L1--L3 (L4 excluded from semantic evaluation; see \S\ref{sec:levels}). ``Entails L0'' means bidirectional NLI entailment against the \emph{same-channel} L0 reference. $C_2$ is correct answers paired with surface-text divergence from the same-channel L0 reference; $C_3$ is reference-text agreement despite an incorrect answer. Accuracy-only evaluation cannot separate these outcomes. Operationalized by bidirectional NLI in the main text and replicated under the alternative criteria of Appendix~\ref{app:semantic_robustness}.}
\label{tab:dissociation}
\end{table}

\subsection{Datasets}
\label{sec:datasets}

We use five datasets spanning four task types and three answer formats (Table~\ref{tab:datasets}): GSM8K \citep{cobbe2021training}, BoolQ \citep{clark2019boolq}, ARC-Easy \citep{clark2018think}, CommonsenseQA \citep{talmor2019commonsenseqa}, and the STEM split of MMLU \citep{hendrycks2020measuring}. BoolQ and CommonsenseQA use validation splits; GSM8K, ARC-Easy, and MMLU-STEM use test splits; MMLU is restricted to 20 STEM subjects.

\begin{table}[!htbp]
\centering
\small
\begin{tabular}{lrl}
\toprule
Dataset & $n$ & Answer type \\
\midrule
GSM8K         & 1{,}319 & numeric   \\
BoolQ         & 3{,}270 & boolean   \\
ARC-Easy      & 2{,}376 & MC (A--D) \\
CommonsenseQA & 1{,}221 & MC (A--E) \\
MMLU-STEM     & 3{,}279 & MC (A--D) \\
\bottomrule
\end{tabular}
\caption{Datasets used in \textsc{Cavewoman}, spanning math word problems (GSM8K), passage yes/no (BoolQ), science multiple-choice (ARC-Easy), commonsense multiple-choice (CommonsenseQA), and STEM multiple-choice (MMLU-STEM). A single model is evaluated on 11{,}465 items at five reduction levels under both conditions.}
\label{tab:datasets}
\end{table}

\subsection{Models}
\label{sec:models}

We evaluate Qwen2.5-VL-7B \citep{bai2025qwen25vl}, Qwen3.5-9B \citep{yang2025qwen3}, DeepSeek-R1-Distill-Qwen-7B \citep{guo2025deepseek}, Gemma-4-E4B \citep{google2026gemma4}, GPT-4o \citep{hurst2024gpt}, GPT-5.4 \citep{openai2026gpt54}, Claude Haiku~4.5 \citep{anthropic2025haikusystemcard}, and Claude Sonnet~4.6 \citep{anthropic2026sonnetsystemcard}. All eight models are evaluated on every benchmark under both conditions. Qwen2.5-VL-7B is included on these text-only benchmarks because the evaluation uses its text backbone in ordinary chat mode rather than any vision input path. Two reasoning-protocol details affect interpretation: DeepSeek-R1 emits hidden \texttt{<think>} traces that count against the same token budget as its visible output, and Qwen3.5-9B has an optional thinking mode that we leave off (the model default). We nonetheless group Qwen3.5-9B with the reasoning models because, like the distilled reasoner, its unconstrained generations are already short and terse, the property that governs how much surface text can diverge under output compression. Mechanics and the exclusion of a third reasoning model (Kimi-K2.6 \cite{kimik262026}) are in Appendix~\ref{app:reasoning_tokens}; full inference configuration is in Appendix~\ref{app:inference_config}.

\section{Results}
\label{sec:results}

\begin{figure*}[t]
\centering
\includegraphics[width=\linewidth]{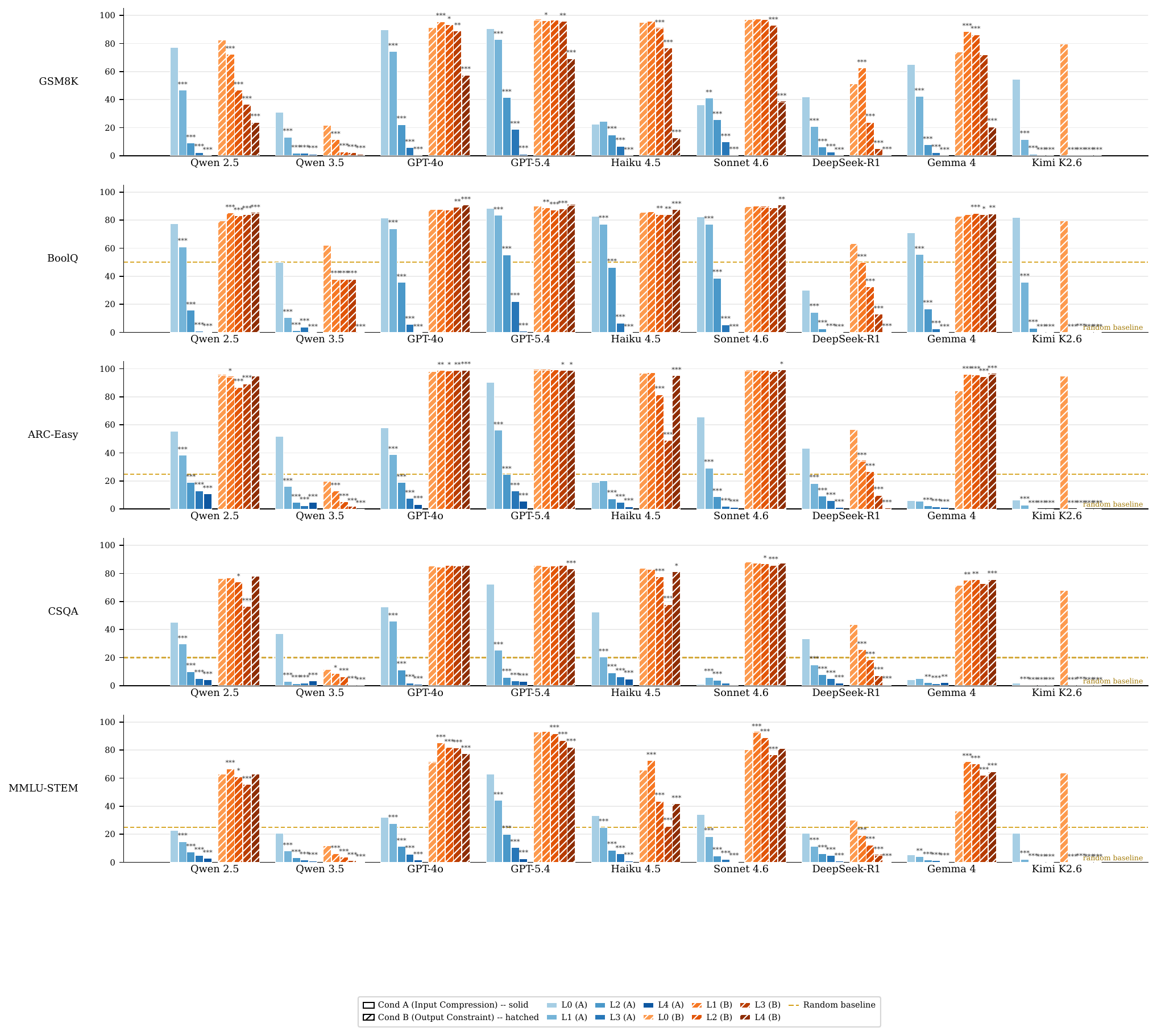}
\caption{Answer accuracy across the five reduction levels for all models and benchmarks. Solid bars denote input compression; hatched bars denote output compression. Significance markers indicate Wilcoxon signed-rank tests against the within-model unconstrained baseline ($^*p{<}.05$, $^{**}p{<}.01$, $^{***}p{<}.001$). The dashed gold line marks the random-guessing baseline for each benchmark. Kimi-K2.6 is shown for completeness only and is excluded from all aggregates (Appendix~E).}
\label{fig:accuracy}
\end{figure*}

\subsection{Finding 1: Cost asymmetry between channels}
\label{sec:finding-cost}

\begin{tcolorbox}[colback=blue!6, colframe=blue!50!black, boxrule=0.6pt, arc=2pt, left=6pt, right=6pt, top=4pt, bottom=4pt]
\textbf{Finding 1.} Output compression cuts realized cost on most API models (\textbf{1.4--2.4$\times$} per model, up to \textbf{3$\times$} on the best cell; cheaper on 17 of 20 `(model, dataset)' cells) and all four open-weight models. Input compression instead raises net cost through compensatory output expansion: up to \textbf{1.15$\times$} on the five-benchmark mean and \textbf{1.8$\times$} on the worst individual dataset, growing to \textbf{2.7$\times$} at deeper reductions as accuracy collapses.
\end{tcolorbox}

\begin{figure*}[t]
\centering
\includegraphics[width=0.92\textwidth]{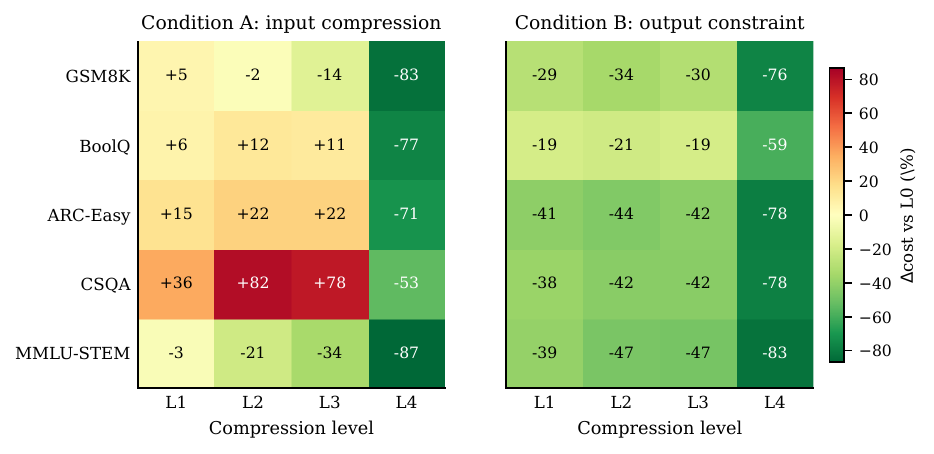}
\caption{Relative change in estimated per-item inference cost against the unconstrained baseline, averaged across the four API models. Left panel: input compression. Right panel: output compression. Rows are benchmarks; columns are the four non-zero reduction levels. Red denotes a cost increase, green a cost reduction. The two channels move in opposite directions at the same reduction level. Worst-case input-channel penalties reach $1.8\times$ at L1 and $\sim 2.7\times$ at deeper reductions (Finding~1).}
\label{fig:channel_cost}
\end{figure*}

Per-item cost for an API-served model is $C = n_{\text{in}}\,p_{\text{in}} + n_{\text{out}}\,p_{\text{out}}$, where $n_{\text{in}}$ and $n_{\text{out}}$ are the input and output token counts and $p_{\text{in}}$, $p_{\text{out}}$ the corresponding per-token prices (May~2026). Figure~\ref{fig:channel_cost} reports the relative change in realized cost against the same-channel unconstrained baseline. Output compression reduces realized per-item cost on GPT-4o, Claude Haiku~4.5, and Claude Sonnet~4.6 by the per-model margins above, cheaper on every one of their fifteen benchmarks; the exception is GPT-5.4, whose billed output is dominated by hidden reasoning tokens (Appendix~\ref{app:reasoning_tokens}), so it is cheaper on only two of its five benchmarks (17 of 20 API cells in all). Input compression at the same reduction levels raises net cost on the remaining models (GPT-4o aside) before accuracy collapses at the strictest level (Figure~\ref{fig:accuracy}). This is a strict lose-lose: the telegraphic level already raises net cost (up to 1.8$\times$ on individual datasets) while degrading accuracy, and at deeper reduction levels the worst-case penalty grows to $\sim$2.7$\times$ as accuracy collapses to single digits; the input channel raises cost and lowers accuracy at the same time.

\paragraph{The mechanism is compensatory output expansion.}
Stripping function words from the prompt saves a small number of input tokens, but the model answers at greater length and output tokens cost several times more than input tokens; the net change is positive. Every API model saves input tokens but spends more on output, and the priced ratio between the two leaves net cost higher on every API model but GPT-4o, where the two effects roughly cancel (Table~\ref{tab:cost_economics}). Output compression instead saves on the output side, which dominates cost: the most favorable case (GPT-4o on ARC-Easy) is roughly three times cheaper at the same accuracy, and on GSM8K the same model trades a small accuracy gain against a halving of cost. The saving requires that the priced output actually shrink under the constraint, which holds for every model whose billed output matches its visible response.

\begin{table}[t]
\centering
\footnotesize
\setlength{\tabcolsep}{4pt}
\begin{tabular}{lrrr}
\toprule
\textbf{Model} & $\Delta$\textbf{in tok} & $\Delta$\textbf{out tok} & \textbf{Net cost} \\
\midrule
GPT-4o            & $-15.3$ & $+0.1$  & $-2.6\%$  \\
GPT-5.4           & $-15.3$ & $+18.0$ & $+15.4\%$ \\
Claude Haiku~4.5  & $-15.7$ & $+9.9$  & $+3.1\%$  \\
Claude Sonnet~4.6 & $-15.7$ & $+12.9$ & $+5.1\%$  \\
\bottomrule
\end{tabular}
\caption{Per-model token economics under input compression at the telegraphic level, averaged across the five benchmarks. Input tokens fall on every model; output tokens rise on every model; net cost is positive on all but one configuration.}
\label{tab:cost_economics}
\end{table}

\paragraph{Apparent accuracy gains can be a parser artifact.}
Answer-extraction rates can inflate apparent accuracy gains under compression. MMLU-STEM provides the clearest example, several API models have lower answer-extraction rates under the unconstrained setting than under compressed output, so apparent improvements partly reflect easier parsing rather than better reasoning. To avoid attributing parser recovery to the model itself, we only report ``compressed exceeds unconstrained'' gains when the unconstrained extraction rate is at least 0.95, and flag the remaining cases as extractor-suppressed (Appendix~\ref{app:extraction_audit}).
\paragraph{Scope of the cost win.}
Output-channel cost savings are largest on benchmarks whose ground-truth answer is already short (BoolQ yes/no, MCQ-letter): the L1 instruction collapses the response to roughly the answer plus minimal reasoning, and savings are partly mechanical. The per-model savings (mean $1.4$--$2.4\times$ on the three models that save) should therefore be read against this format-collapse component; on GSM8K, where the answer requires multi-step arithmetic, savings are smaller but accuracy is preserved on the cells with $\geq 0.95$ L0 parse rate. Apparent single-cell gains on MMLU-STEM (e.g.\ Gemma-4-E4B, $+35.3$~pp) are confounded by answer-extraction recovery (the L0-B parse rate rises from $0.49$ to $0.88$ at L1) and are reported as extractor-suppressed rather than reasoning gains (Appendix~\ref{app:extraction_audit}).

Under public-tier pricing, all four open-weight models also save cost at L1 under output compression, in the same direction as the API panel (Appendix~\ref{app:open_weight_projection}).

\subsubsection{Channel-specific degradation}
\label{sec:finding-cost-degradation}

The two channels also degrade differently. Under input compression, accuracy and reference-text agreement fall together as the level increases; under output compression, classification accuracy holds through deep levels while reference-text agreement falls sharply at the first level. The input channel spends accuracy; the output channel spends agreement with the unconstrained reference. The output-channel pattern motivates Finding~2.

\subsection{Finding 2: Accuracy decouples from same-channel reference text under output compression}
\label{sec:finding-dissociation}

\begin{tcolorbox}[colback=blue!6, colframe=blue!50!black, boxrule=0.6pt, arc=2pt, left=6pt, right=6pt, top=4pt, bottom=4pt]
\textbf{Finding 2.} Across the six non-reasoning models, $\mathbf{51.9\%}$ of all L1 output-compression generations are correct yet have surface text that no longer matches what the model would have written without the constraint on the six-non-reasoning panel; under length-matched re-scoring on the same panel, the rate rises to $\mathbf{80.4\%}$.
\end{tcolorbox}

The headline rate is the share of all generations at L1 output compression that are correct yet no longer entail the same-channel unconstrained reference under a bidirectional NLI judge (the $C_2$ cell of Table~\ref{tab:dissociation}). Pooled across the six non-reasoning models in our panel, the rate is 51.9\% on the six-non-reasoning panel. The two reasoning models in the panel (DeepSeek-R1-Distill and Qwen3.5-9B) show a smaller divergence since their unconstrained generations are already short (per-model values in Appendix~\ref{app:noise_floor}). On every non-reasoning model, the dominant off-diagonal outcome is correct-but-divergent rather than incorrect-but-faithful.

\begin{figure}[t]
\centering
\includegraphics[width=\columnwidth]{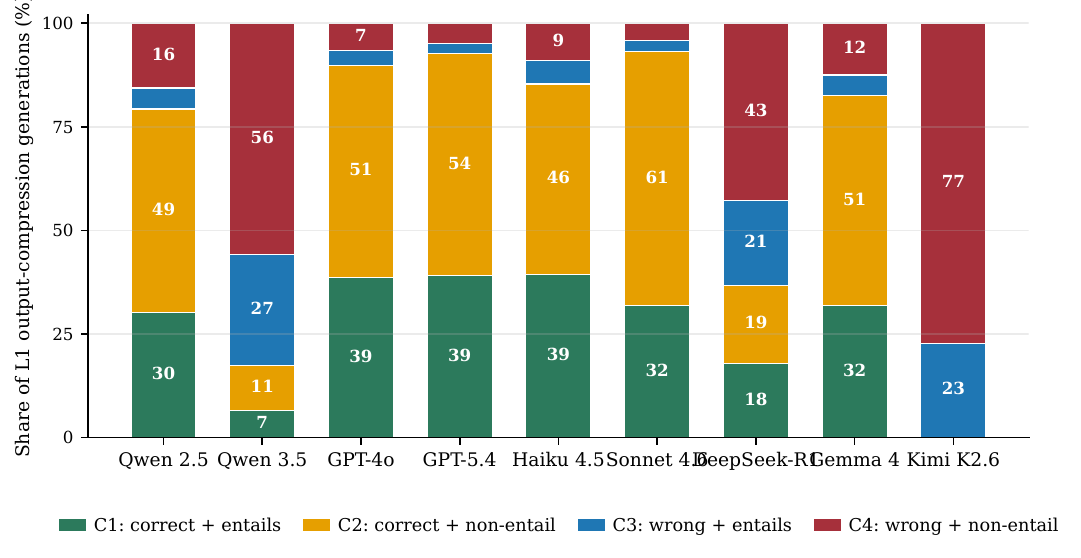}
\caption{Per-model $2{\times}2$ dissociation at L1 output compression, summed across the five benchmarks. Each stacked bar partitions L1 output-compression generations into the four outcome cells: $C_1$ (correct + entails L0), $C_2$ (correct + does not entail L0), $C_3$ (wrong + entails L0), and $C_4$ (wrong + does not entail L0). Accuracy is $C_1{+}C_2$ and bidirectional NLI rate is $C_1{+}C_3$; the amber $C_2$ share is the dissociation. DeepSeek-R1 inverts the pattern. Kimi-K2.6 is shown for completeness only and is excluded from all aggregates (Appendix~E).}
\label{fig:accuracy_vs_semantic}
\end{figure}

\paragraph{Accuracy holds while the reasoning trace drifts.}
Figure~\ref{fig:accuracy_vs_semantic} resolves L1 output compression into its $2{\times}2$ outcome cells per model. The amber $C_2$ segment is the dissociation: correct answers whose surface text no longer entails the model's same-channel L0 reference. On every non-reasoning model the $C_2$ band is the dominant off-diagonal cell; DeepSeek-R1 inverts the pattern (Appendix~\ref{app:reasoning_tokens}).

\paragraph{Length is not the explanation.}
Length-matched re-scoring (truncating L0 to the L1-B wordpiece-token length) \emph{increases} the divergence on every non-reasoning model (Table~\ref{tab:length_controlled}, Appendix~\ref{app:noise_floor}); the headline is the conservative reading.

\paragraph{Robustness across judges, metrics, and statistics.}
The bidirectional NLI judge has a calibrated false-negative rate of $2.9\%$ at L1 on synthetic positive pairs (Appendix~\ref{app:judge_reliability}); all twelve semantic measures report a substantial divergence, with our headline sitting near the conservative end of the family (Table~\ref{tab:alt_metrics}). Wilcoxon signed-rank tests against the within-model, within-channel unconstrained baseline survive Benjamini--Hochberg correction at $\alpha=0.05$ on every significant cell (Appendix~\ref{app:limitations_detail}).

\paragraph{Generalization and task-type dependence.}
The divergence also holds under the learned LLMLingua-2 compressor \citep{pan2024llmlingua}: $C_2 > 0$ on every cell at $\tau{=}0.5$, with rate-driven recovery at $\tau{=}0.8$ (Appendix~\ref{app:llmlingua}). The divergence is smaller on arithmetic benchmarks than on classification (Table~\ref{tab:task_comparison}). Register traces across L1--L4 are in Appendix~\ref{app:examples}.

\subsection{Robustness to compression varies across models}
\label{sec:finding-modelgroup}

\begin{tcolorbox}[colback=blue!6, colframe=blue!50!black, boxrule=0.6pt, arc=2pt, left=6pt, right=6pt, top=4pt, bottom=4pt]
\textbf{Finding 3.} Relative robustness (telegraphic output- vs.\ input-constraint accuracy) spans a $\mathbf{1.5\times}$ to $\mathbf{3.7\times}$ range across models and is not predicted by parameter count or unconstrained accuracy.
\end{tcolorbox}

Let robustness be the ratio of accuracy under the telegraphic output constraint to accuracy under the telegraphic input constraint, at the same item and the same reduction level. The ratio is reported per model in Table~\ref{tab:robustness}. The most output-robust model overall is the open-weight Gemma-4-E4B, and the most output-robust API model (Claude Sonnet~4.6) retains nearly three times as much accuracy under the output constraint as under the input constraint; the highest-accuracy API model on the unconstrained baseline (GPT-5.4) is least robust among the four API models (second-least across all eight). Gemma-4-E4B tops the full ranking, and Qwen2.5-VL-7B outranks two of the four API models. DeepSeek-R1's value should be read with the reasoning-token caveat of Appendix~\ref{app:reasoning_tokens}, since its hidden chain-of-thought tokens are charged against the same visible-output budget; with two API vendors at two models each, we draw no vendor-level conclusion. The practical implication is that a deployment should rank candidates at the constraint level under which it will be deployed, not at the unconstrained baseline.

\begin{table}[!htbp]
\centering
\small
\setlength{\tabcolsep}{6pt}
\begin{tabular}{lc}
\toprule
\textbf{Model} & \textbf{Output / input acc. ratio} \\
\midrule
Gemma-4-E4B       & 3.70 \\
Claude Sonnet~4.6 & 2.71 \\
Claude Haiku~4.5  & 2.60 \\
DeepSeek-R1       & 2.40 \\
Qwen2.5-VL-7B     & 2.07 \\
GPT-4o            & 1.73 \\
GPT-5.4           & 1.59 \\
Qwen3.5-9B        & 1.50 \\
\bottomrule
\end{tabular}
\caption{Output-vs-input accuracy ratio at the telegraphic level, per model. Higher values indicate greater robustness to output compression. Parameter count does not predict the ordering.}
\label{tab:robustness}
\end{table}

\section{Discussion}
\label{sec:discussion}

Output is the channel to compress on when the model answers at length (Finding~1). It produces real cost savings on every API model whose billed output matches its visible response, while input compression does not.

Robustness to output compression does not follow from unconstrained accuracy. The highest-accuracy API model in our panel is least robust among the four API models (second-least across all eight), and a 7B open-weight model outranks two of the four API models under the output constraint.

The accuracy/reference-text divergence is a surface-text observation, not a propositional claim. Findings~1 and~2 answer different questions and can both hold: correctness is graded against the ground-truth answer, while reference-text agreement is graded against the model's own L0 generation. For deployments that consume only the final answer, the divergence does not affect outcomes; for deployments that consume the generation as text (transcripts, audit trails, reasoning displays), it does.

These findings suggest that single-axis evaluation of compression is underdetermined. Realised cost on the priced channel is not reducible to prompt-token reduction; observed accuracy is conditional on extraction-rate reliability; and reference-text agreement is conditional on the choice of semantic axis. The three quantities dissociate in our panel; any composite metric that aggregates them can therefore mis-rank methods on the dimension that ultimately determines deployment cost.

\section{Conclusion}
\label{sec:conclusion}

Compressing language-model inference is a two-channel problem, and accuracy alone cannot tell the channels apart. Output compression cuts realized cost on most API models (1.4--2.4$\times$ per model, up to 3$\times$ on the best cell, at the first reduction level) and all four open-weight models under public-tier pricing; input compression instead raises net cost through compensatory output expansion (up to $\sim$15\% on the five-benchmark mean and 1.8$\times$ on individual datasets, growing to 2.7$\times$ at deeper reductions as accuracy collapses). On the same cost-saving settings, 51.9\% of generations on the six-non-reasoning panel are correct yet their surface text no longer matches the model's unconstrained reference, a divergence that strengthens under length-controlled re-scoring and replicates across complementary semantic measures. Robustness to output compression varies widely across models and is not predicted by parameter count or unconstrained accuracy; candidates should therefore be ranked at the constraint level they will be deployed at, not at the unconstrained baseline.

\subsection{Limitations}
\label{sec:limitations}

Our bidirectional NLI judge measures surface-text divergence rather than propositional drift; we bound the length-and-register confound through length-controlled re-scoring (Appendix~\ref{app:noise_floor}) and replicate under eleven complementary measures plus the headline judge (Appendix~\ref{app:semantic_robustness}), but the rate is not a propositional-content claim. Part of the Condition~B divergence may also reflect the system-prompt register change between conditions; the L0-A/L0-B noise floor that would isolate this is not separately measured (Appendix~\ref{app:noise_floor}). All five benchmarks have short, structured answers (numeric, boolean, MCQ letter); we make no claim about content preservation in long-form generation tasks such as summarization or open-ended QA. The eight-model panel uses greedy decoding only, with two API vendors at two models each; we therefore draw no vendor- or family-level conclusion; sampled decoding and a hard-decoder L4 are out of scope. Full per-confound discussion is in Appendix~\ref{app:limitations_detail}.

\section*{Acknowledgments}

We thank MIT Engaging for providing the GPU compute used for the local-model runs.

\bibliography{custom}

@inproceedings{jiang-etal-2023-llmlingua,
    title = "{LLML}ingua: Compressing Prompts for Accelerated Inference of Large Language Models",
    author = "Jiang, Huiqiang  and
      Wu, Qianhui  and
      Lin, Chin-Yew  and
      Yang, Yuqing  and
      Qiu, Lili",
    editor = "Bouamor, Houda  and
      Pino, Juan  and
      Bali, Kalika",
    booktitle = "Proceedings of the 2023 Conference on Empirical Methods in Natural Language Processing",
    month = dec,
    year = "2023",
    address = "Singapore",
    publisher = "Association for Computational Linguistics",
    url = "https://aclanthology.org/2023.emnlp-main.825/",
    doi = "10.18653/v1/2023.emnlp-main.825",
    pages = "13358--13376"
}

@inproceedings{li2023compressing,
  title={Compressing context to enhance inference efficiency of large language models},
  author={Li, Yucheng and Dong, Bo and Guerin, Frank and Lin, Chenghua},
  booktitle={Proceedings of the 2023 conference on empirical methods in natural language processing},
  pages={6342--6353},
  year={2023}
}

@inproceedings{pan2024llmlingua,
  title={Llmlingua-2: Data distillation for efficient and faithful task-agnostic prompt compression},
  author={Pan, Zhuoshi and Wu, Qianhui and Jiang, Huiqiang and Xia, Menglin and Luo, Xufang and Zhang, Jue and Lin, Qingwei and R{\"u}hle, Victor and Yang, Yuqing and Lin, Chin-Yew and others},
  booktitle={Findings of the Association for Computational Linguistics: ACL 2024},
  pages={963--981},
  year={2024}
}

@inproceedings{jiang2024longllmlingua,
  title={Longllmlingua: Accelerating and enhancing llms in long context scenarios via prompt compression},
  author={Jiang, Huiqiang and Wu, Qianhui and Luo, Xufang and Li, Dongsheng and Lin, Chin-Yew and Yang, Yuqing and Qiu, Lili},
  booktitle={Proceedings of the 62nd Annual Meeting of the Association for Computational Linguistics (Volume 1: Long Papers)},
  pages={1658--1677},
  year={2024}
}

@inproceedings{xu2024recomp,
  title={Recomp: Improving retrieval-augmented lms with context compression and selective augmentation},
  author={Xu, Fangyuan and Shi, Weijia and Choi, Eunsol},
  booktitle={International Conference on Learning Representations},
  volume={2024},
  pages={43478--43502},
  year={2024}
}

@article{mu2023learning,
  title={Learning to compress prompts with gist tokens},
  author={Mu, Jesse and Li, Xiang and Goodman, Noah},
  journal={Advances in Neural Information Processing Systems},
  volume={36},
  pages={19327--19352},
  year={2023}
}

@inproceedings{chevalier2023adapting,
  title={Adapting language models to compress contexts},
  author={Chevalier, Alexis and Wettig, Alexander and Ajith, Anirudh and Chen, Danqi},
  booktitle={Proceedings of the 2023 Conference on Empirical Methods in Natural Language Processing},
  pages={3829--3846},
  year={2023}
}

@inproceedings{
ge2024incontext,
title={In-context Autoencoder for Context Compression in a Large Language Model},
author={Tao Ge and Hu Jing and Lei Wang and Xun Wang and Si-Qing Chen and Furu Wei},
booktitle={The Twelfth International Conference on Learning Representations},
year={2024},
url={https://openreview.net/forum?id=uREj4ZuGJE}
}

@inproceedings{li2025prompt,
  title={Prompt compression for large language models: A survey},
  author={Li, Zongqian and Liu, Yinhong and Su, Yixuan and Collier, Nigel},
  booktitle={Proceedings of the 2025 Conference of the Nations of the Americas Chapter of the Association for Computational Linguistics: Human Language Technologies (Volume 1: Long Papers)},
  pages={7182--7195},
  year={2025}
}

@article{nagle2024fundamental,
  title={Fundamental limits of prompt compression: A rate-distortion framework for black-box language models},
  author={Nagle, Alliot and Girish, Adway and Bondaschi, Marco and Gastpar, Michael and Makkuva, Ashok Vardhan and Kim, Hyeji},
  journal={Advances in Neural Information Processing Systems},
  volume={37},
  pages={94934--94970},
  year={2024}
}

@inproceedings{kikuchi2016controlling,
  title={Controlling output length in neural encoder-decoders},
  author={Kikuchi, Yuta and Neubig, Graham and Sasano, Ryohei and Takamura, Hiroya and Okumura, Manabu},
  booktitle={Proceedings of the 2016 Conference on Empirical Methods in Natural Language Processing},
  pages={1328--1338},
  year={2016}
}

@inproceedings{takase-okazaki-2019-positional,
    title = "Positional Encoding to Control Output Sequence Length",
    author = "Takase, Sho  and
      Okazaki, Naoaki",
    editor = "Burstein, Jill  and
      Doran, Christy  and
      Solorio, Thamar",
    booktitle = "Proceedings of the 2019 Conference of the North {A}merican Chapter of the Association for Computational Linguistics: Human Language Technologies, Volume 1 (Long and Short Papers)",
    month = jun,
    year = "2019",
    address = "Minneapolis, Minnesota",
    publisher = "Association for Computational Linguistics",
    url = "https://aclanthology.org/N19-1401/",
    doi = "10.18653/v1/N19-1401",
    pages = "3999--4004"
}

@inproceedings{song2025hansel,
  title={Hansel: Output length controlling framework for large language models},
  author={Song, Seoha and Lee, Junhyun and Ko, Hyeonmok},
  booktitle={Proceedings of the AAAI Conference on Artificial Intelligence},
  number={23},
  pages={25146--25154},
  year={2025}
}

@inproceedings{xia2025tokenskip,
  title={Tokenskip: Controllable chain-of-thought compression in llms},
  author={Xia, Heming and Leong, Chak Tou and Wang, Wenjie and Li, Yongqi and Li, Wenjie},
  booktitle={Proceedings of the 2025 Conference on Empirical Methods in Natural Language Processing},
  pages={3351--3363},
  year={2025}
}

@inproceedings{aytes2025sketch,
  title={Sketch-of-thought: Efficient llm reasoning with adaptive cognitive-inspired sketching},
  author={Aytes, Simon A and Baek, Jinheon and Hwang, Sung Ju},
  booktitle={Proceedings of the 2025 Conference on Empirical Methods in Natural Language Processing},
  pages={24307--24331},
  year={2025}
}

@inproceedings{han2025token,
  title={Token-budget-aware llm reasoning},
  author={Han, Tingxu and Wang, Zhenting and Fang, Chunrong and Zhao, Shiyu and Ma, Shiqing and Chen, Zhenyu},
  booktitle={Findings of the Association for Computational Linguistics: ACL 2025},
  pages={24842--24855},
  year={2025}
}

@inproceedings{huang2024fewer,
  title={Fewer is more: Boosting math reasoning with reinforced context pruning},
  author={Huang, Xijie and Zhang, Li Lyna and Cheng, Kwang-Ting and Yang, Fan and Yang, Mao},
  booktitle={Proceedings of the 2024 Conference on Empirical Methods in Natural Language Processing},
  pages={13674--13695},
  year={2024}
}

@inproceedings{
kuhn2023semantic,
title={Semantic Uncertainty: Linguistic Invariances for Uncertainty Estimation in Natural Language Generation},
author={Lorenz Kuhn and Yarin Gal and Sebastian Farquhar},
booktitle={The Eleventh International Conference on Learning Representations },
year={2023},
url={https://openreview.net/forum?id=VD-AYtP0dve}
}

@inproceedings{maynez2020faithfulness,
  title={On faithfulness and factuality in abstractive summarization},
  author={Maynez, Joshua and Narayan, Shashi and Bohnet, Bernd and McDonald, Ryan},
  booktitle={Proceedings of the 58th annual meeting of the association for computational linguistics},
  pages={1906--1919},
  year={2020}
}

@article{laban2022summac,
  title={SummaC: Re-visiting NLI-based models for inconsistency detection in summarization},
  author={Laban, Philippe and Schnabel, Tobias and Bennett, Paul N and Hearst, Marti A},
  journal={Transactions of the Association for Computational Linguistics},
  volume={10},
  pages={163--177},
  year={2022},
  publisher={MIT Press One Rogers Street, Cambridge, MA 02142-1209, USA journals-info~…}
}

@article{turpin2023language,
  title={Language models don't always say what they think: Unfaithful explanations in chain-of-thought prompting},
  author={Turpin, Miles and Michael, Julian and Perez, Ethan and Bowman, Samuel},
  journal={Advances in Neural Information Processing Systems},
  volume={36},
  pages={74952--74965},
  year={2023}
}

@article{lanham2023measuring,
  title={Measuring faithfulness in chain-of-thought reasoning},
  author={Lanham, Tamera and Chen, Anna and Radhakrishnan, Ansh and Steiner, Benoit and Denison, Carson and Hernandez, Danny and Li, Dustin and Durmus, Esin and Hubinger, Evan and Kernion, Jackson and others},
  journal={arXiv preprint arXiv:2307.13702},
  year={2023}
}

@inproceedings{zhang2025demystify,
  title={Demystify Verbosity Compensation Behavior of Large Language Models},
  author={Zhang, Yusen and Das, Sarkar Snigdha Sarathi and Zhang, Rui},
  booktitle={Proceedings of the 2nd Workshop on Uncertainty-Aware NLP (UncertaiNLP 2025)},
  pages={160--178},
  year={2025}
}

@article{borisov2026chatbot,
  title={Do Chatbot LLMs Talk Too Much? The YapBench Benchmark},
  author={Borisov, Vadim and Gr{\"o}ger, Michael and Mikhael, Mina and Schreiber, Richard H},
  journal={arXiv preprint arXiv:2601.00624},
  year={2026}
}

@inproceedings{levy2024same,
  title={Same task, more tokens: the impact of input length on the reasoning performance of large language models},
  author={Levy, Mosh and Jacoby, Alon and Goldberg, Yoav},
  booktitle={Proceedings of the 62nd Annual Meeting of the Association for Computational Linguistics (Volume 1: Long Papers)},
  pages={15339--15353},
  year={2024}
}

@inproceedings{jin2024impact,
  title={The impact of reasoning step length on large language models},
  author={Jin, Mingyu and Yu, Qinkai and Shu, Dong and Zhao, Haiyan and Hua, Wenyue and Meng, Yanda and Zhang, Yongfeng and Du, Mengnan},
  booktitle={Findings of the Association for Computational Linguistics: ACL 2024},
  pages={1830--1842},
  year={2024}
}

@inproceedings{sun2025empirical,
  title={An empirical study of llm reasoning ability under strict output length constraint},
  author={Sun, Yi and Wang, Han and Li, Jiaqiang and Liu, Jiacheng and Li, Xiangyu and Wen, Hao and Yuan, Yizhen and Zheng, Huiwen and Liang, Yan and Li, Yuanchun and others},
  booktitle={Proceedings of the 2025 Conference on Empirical Methods in Natural Language Processing},
  pages={7663--7682},
  year={2025}
}

@inproceedings{wang2024reasoning,
  title={Reasoning in token economies: budget-aware evaluation of LLM reasoning strategies},
  author={Wang, Junlin and Jain, Siddhartha and Zhang, Dejiao and Ray, Baishakhi and Kumar, Varun and Athiwaratkun, Ben},
  booktitle={Proceedings of the 2024 Conference on Empirical Methods in Natural Language Processing},
  pages={19916--19939},
  year={2024}
}

@article{
chen2024frugalgpt,
title={Frugal{GPT}: How to Use Large Language Models While Reducing Cost and Improving Performance},
author={Lingjiao Chen and Matei Zaharia and James Zou},
journal={Transactions on Machine Learning Research},
issn={2835-8856},
year={2024},
url={https://openreview.net/forum?id=cSimKw5p6R},
note={Featured Certification}
}

@inproceedings{ding2024hybrid,
  title={Hybrid llm: Cost-efficient and quality-aware query routing},
  author={Ding, Dujian and Mallick, Ankur and Wang, Chi and Sim, Robert and Mukherjee, Subhabrata and R{\"u}hle, Victor and Lakshmanan, Laks and Awadallah, Ahmed H},
  booktitle={International Conference on Learning Representations},
  volume={2024},
  pages={41348--41366},
  year={2024}
}

@inproceedings{
ong2025routellm,
title={Route{LLM}: Learning to Route {LLM}s from Preference Data},
author={Isaac Ong and Amjad Almahairi and Vincent Wu and Wei-Lin Chiang and Tianhao Wu and Joseph E. Gonzalez and M Waleed Kadous and Ion Stoica},
booktitle={The Thirteenth International Conference on Learning Representations},
year={2025},
url={https://openreview.net/forum?id=8sSqNntaMr}
}

@inproceedings{bowman2021will,
  title={What will it take to fix benchmarking in natural language understanding?},
  author={Bowman, Samuel and Dahl, George},
  booktitle={Proceedings of the 2021 Conference of the North American Chapter of the Association for Computational Linguistics: Human Language Technologies},
  pages={4843--4855},
  year={2021}
}

@article{
liang2023holistic,
title={Holistic Evaluation of Language Models},
author={Percy Liang and Rishi Bommasani and Tony Lee and Dimitris Tsipras and Dilara Soylu and Michihiro Yasunaga and Yian Zhang and Deepak Narayanan and Yuhuai Wu and Ananya Kumar and Benjamin Newman and Binhang Yuan and Bobby Yan and Ce Zhang and Christian Cosgrove and Christopher D Manning and Christopher Re and Diana Acosta-Navas and Drew A. Hudson and Eric Zelikman and Esin Durmus and Faisal Ladhak and Frieda Rong and Hongyu Ren and Huaxiu Yao and Jue WANG and Keshav Santhanam and Laurel Orr and Lucia Zheng and Mert Yuksekgonul and Mirac Suzgun and Nathan Kim and Neel Guha and Niladri S. Chatterji and Omar Khattab and Peter Henderson and Qian Huang and Ryan Andrew Chi and Sang Michael Xie and Shibani Santurkar and Surya Ganguli and Tatsunori Hashimoto and Thomas Icard and Tianyi Zhang and Vishrav Chaudhary and William Wang and Xuechen Li and Yifan Mai and Yuhui Zhang and Yuta Koreeda},
journal={Transactions on Machine Learning Research},
issn={2835-8856},
year={2023},
url={https://openreview.net/forum?id=iO4LZibEqW},
note={Featured Certification, Expert Certification, Outstanding Certification}
}

@inproceedings{gehrmann2021gem,
  title={The gem benchmark: Natural language generation, its evaluation and metrics},
  author={Gehrmann, Sebastian and Adewumi, Tosin and Aggarwal, Karmanya and Ammanamanchi, Pawan Sasanka and Aremu, Anuoluwapo and Bosselut, Antoine and Chandu, Khyathi Raghavi and Clinciu, Miruna and Das, Dipanjan and Dhole, Kaustubh and others},
  booktitle={Proceedings of the 1st Workshop on Natural Language Generation, Evaluation, and Metrics (GEM 2021)},
  pages={96--120},
  year={2021}
}

@article{cobbe2021training,
  title={Training verifiers to solve math word problems},
  author={Cobbe, Karl and Kosaraju, Vineet and Bavarian, Mohammad and Chen, Mark and Jun, Heewoo and Kaiser, Lukasz and Plappert, Matthias and Tworek, Jerry and Hilton, Jacob and Nakano, Reiichiro and others},
  journal={arXiv preprint arXiv:2110.14168},
  year={2021}
}

@inproceedings{clark2019boolq,
  title={Boolq: Exploring the surprising difficulty of natural yes/no questions},
  author={Clark, Christopher and Lee, Kenton and Chang, Ming-Wei and Kwiatkowski, Tom and Collins, Michael and Toutanova, Kristina},
  booktitle={Proceedings of the 2019 conference of the north American chapter of the association for computational linguistics: Human language technologies, volume 1 (long and short papers)},
  pages={2924--2936},
  year={2019}
}

@article{clark2018think,
  title={Think you have solved question answering? try arc, the ai2 reasoning challenge},
  author={Clark, Peter and Cowhey, Isaac and Etzioni, Oren and Khot, Tushar and Sabharwal, Ashish and Schoenick, Carissa and Tafjord, Oyvind},
  journal={arXiv preprint arXiv:1803.05457},
  year={2018}
}

@inproceedings{talmor2019commonsenseqa,
  title={Commonsenseqa: A question answering challenge targeting commonsense knowledge},
  author={Talmor, Alon and Herzig, Jonathan and Lourie, Nicholas and Berant, Jonathan},
  booktitle={Proceedings of the 2019 Conference of the North American Chapter of the Association for Computational Linguistics: Human Language Technologies, Volume 1 (Long and Short Papers)},
  pages={4149--4158},
  year={2019}
}

@inproceedings{
hendrycks2020measuring,
title={Measuring Massive Multitask Language Understanding},
author={Dan Hendrycks and Collin Burns and Steven Basart and Andy Zou and Mantas Mazeika and Dawn Song and Jacob Steinhardt},
booktitle={International Conference on Learning Representations},
year={2021},
url={https://openreview.net/forum?id=d7KBjmI3GmQ}
}

@inproceedings{
he2020deberta,
title={{\{}DEBERTA{\}}: {\{}DECODING{\}}-{\{}ENHANCED{\}} {\{}BERT{\}} {\{}WITH{\}} {\{}DISENTANGLED{\}} {\{}ATTENTION{\}}},
author={Pengcheng He and Xiaodong Liu and Jianfeng Gao and Weizhu Chen},
booktitle={International Conference on Learning Representations},
year={2021},
url={https://openreview.net/forum?id=XPZIaotutsD}
}

@inproceedings{reimers-gurevych-2019-sentence,
    title = "Sentence-{BERT}: Sentence Embeddings using {S}iamese {BERT}-Networks",
    author = "Reimers, Nils  and
      Gurevych, Iryna",
    editor = "Inui, Kentaro  and
      Jiang, Jing  and
      Ng, Vincent  and
      Wan, Xiaojun",
    booktitle = "Proceedings of the 2019 Conference on Empirical Methods in Natural Language Processing and the 9th International Joint Conference on Natural Language Processing (EMNLP-IJCNLP)",
    month = nov,
    year = "2019",
    address = "Hong Kong, China",
    publisher = "Association for Computational Linguistics",
    url = "https://aclanthology.org/D19-1410/",
    doi = "10.18653/v1/D19-1410",
    pages = "3982--3992"
}

@article{bai2025qwen25vl,
  title   = {{Qwen2.5-VL} Technical Report},
  author  = {Bai, Shuai and Chen, Keqin and Liu, Xuejing and Wang, Jialin and
             Ge, Wenbin and Song, Sibo and Dang, Kai and Wang, Peng and
             Wang, Shijie and Tang, Jun and Zhong, Humen and Zhu, Yuanzhi
             and Yang, Mingkun and Li, Zhaohai and Wan, Jianqiang and Wang,
             Pengfei and Ding, Wei and Fu, Zheren and Xu, Yiheng and Ye,
             Jiabo and Zhang, Xi and Xie, Tianbao and Cheng, Zesen and Zhang,
             Hang and Yang, Zhibo and Xu, Haiyang and Lin, Junyang},
  journal = {arXiv preprint arXiv:2502.13923},
  year    = {2025},
}

@article{yang2025qwen3,
  title={Qwen3 technical report},
  author={Yang, An and Li, Anfeng and Yang, Baosong and Zhang, Beichen and Hui, Binyuan and Zheng, Bo and Yu, Bowen and Gao, Chang and Huang, Chengen and Lv, Chenxu and others},
  journal={arXiv preprint arXiv:2505.09388},
  year={2025}
}

@misc{google2026gemma4,
  title        = {{Gemma 4}: Our Most Capable Open Models to Date},
  author       = {{Google}},
  year         = {2026},
  howpublished = {\url{https://blog.google/innovation-and-ai/technology/developers-tools/gemma-4/}},
  note         = {Google Blog}
}

@article{guo2025deepseek,
  title={Deepseek-r1: Incentivizing reasoning capability in llms via reinforcement learning},
  author={Guo, Daya and Yang, Dejian and Zhang, Haowei and Song, Junxiao and Wang, Peiyi and Zhu, Qihao and Xu, Runxin and Zhang, Ruoyu and Ma, Shirong and Bi, Xiao and others},
  journal={arXiv preprint arXiv:2501.12948},
  year={2025}
}

@article{hurst2024gpt,
  title={Gpt-4o system card},
  author={Hurst, Aaron and Lerer, Adam and Goucher, Adam P and Perelman, Adam and Ramesh, Aditya and Clark, Aidan and Ostrow, AJ and Welihinda, Akila and Hayes, Alan and Radford, Alec and others},
  journal={arXiv preprint arXiv:2410.21276},
  year={2024}
}

@techreport{anthropic2025haikusystemcard,
  title       = {System Card: Claude Haiku 4.5},
  author      = {{Anthropic}},
  institution = {Anthropic},
  year        = {2025},
  month       = oct,
  note        = {System card documenting improvements and model safety testing for Claude Haiku 4.5},
  url         = {https://www-cdn.anthropic.com/7aad69bf12627d42234e01ee7c36305dc2f6a970.pdf}
}

@misc{openai2026gpt54,
  title        = {Introducing {GPT-5.4}: Designed for Professional Work},
  author       = {{OpenAI}},
  year         = {2026},
  month        = mar,
  howpublished = {\url{https://openai.com/index/introducing-gpt-5-4/}},
  note         = {OpenAI Blog}
}

@techreport{anthropic2026sonnetsystemcard,
  title       = {System Card: Claude Sonnet 4.6},
  author      = {{Anthropic}},
  institution = {Anthropic},
  year        = {2026},
  month       = sep,
  note        = {Technical system card documenting model safeguards, model characteristics, and deployment of Claude Sonnet 4.6},
  url         = {https://www-cdn.anthropic.com/bbd8ef16d70b7a1665f14f306ee88b53f686aa75.pdf}
}

@misc{brussee2026caveman,
  title        = {Caveman},
  author       = {Julius Brussee},
  year         = {2026},
  howpublished = {\url{https://github.com/juliusbrussee/caveman}},
  note         = {GitHub repository}
}

@misc{pelser2026cavemancompression,
  title        = {Caveman Compression},
  author       = {William Peltom\"aki},
  year         = {2026},
  howpublished = {\url{https://github.com/wilpel/caveman-compression}},
  note         = {GitHub repository}
}

@inproceedings{nag-etal-2024-cost,
    title = "Cost-Performance Optimization for Processing Low-Resource Language Tasks Using Commercial {LLM}s",
    author = "Nag, Arijit  and
      Mukherjee, Animesh  and
      Ganguly, Niloy  and
      Chakrabarti, Soumen",
    editor = "Al-Onaizan, Yaser  and
      Bansal, Mohit  and
      Chen, Yun-Nung",
    booktitle = "Findings of the Association for Computational Linguistics: EMNLP 2024",
    month = nov,
    year = "2024",
    address = "Miami, Florida, USA",
    publisher = "Association for Computational Linguistics",
    url = "https://aclanthology.org/2024.findings-emnlp.920/",
    doi = "10.18653/v1/2024.findings-emnlp.920",
    pages = "15681--15701"
}

@inproceedings{
ahia2023do,
title={Do All Languages Cost the Same? Tokenization in the Era of Commercial Language Models},
author={Orevaoghene Ahia and Sachin Kumar and Hila Gonen and Jungo Kasai and David R Mortensen and Noah A. Smith and Yulia Tsvetkov},
booktitle={The 2023 Conference on Empirical Methods in Natural Language Processing},
year={2023},
url={https://openreview.net/forum?id=OUmxBN45Gl}
}

@misc{kimik262026,
  title={Kimi K2.6 Technical Report}, 
  author={{Kimi Team}},
  year={2026},
  publisher={Moonshot AI},
  url={https://platform.kimi.ai/docs/overview}
}

\appendix

\FloatBarrier
\section{Implementation Details}
\label{app:tokencost}

Pricing, dataset statistics and licenses, inference configuration, token accounting, cost estimates, and the L1 decoder-truncation check. Verbatim system prompts and POS-tag rules are in Appendix~\ref{app:condA}; released artifacts in Appendix~\ref{app:reproducibility}.

\subsection{Pricing Assumptions and Model Snapshots}
\label{app:pricing}

May~2026 per-token API prices for the four closed models (Table~\ref{tab:pricing}). Per-item API cost is $C = n_{\text{in}} \cdot p_{\text{in}} + n_{\text{out}} \cdot p_{\text{out}}$, with tokens counted by each model's own tokenizer. Open-weight models run on local GPUs and are excluded from the dollar-cost analysis; token accounting is in Appendix~\ref{app:full}.

\begin{table*}[!htbp]
\centering
\small
\begin{tabular}{lrrl}
\toprule
\textbf{Model} & \textbf{Input} & \textbf{Output} & \textbf{Snapshot / endpoint} \\
 & \multicolumn{2}{c}{\footnotesize(\$/M tokens)} & \\
\midrule
GPT-4o            & 2.50  & 10.00 & \texttt{gpt-4o-2024-11-20} \\
GPT-5.4           & 1.25  & 10.00 & \texttt{gpt-5-2025-08-07} \\
Claude Haiku~4.5  & 1.00  &  5.00 & \texttt{claude-haiku-4-5-20251001} \\
Claude Sonnet~4.6 & 3.00  & 15.00 & \texttt{claude-sonnet-4-6-20250929} \\
\midrule
Qwen~2.5-VL-7B    & \multicolumn{2}{c}{local GPU} & \texttt{Qwen/Qwen2.5-VL-7B-Instruct} \\
Qwen~3.5-9B       & \multicolumn{2}{c}{local GPU} & \texttt{Qwen/Qwen3.5-9B} \\
DeepSeek-R1-Distill 7B & \multicolumn{2}{c}{local GPU} & \texttt{deepseek-ai/DeepSeek-R1-Distill-Qwen-7B} \\
Gemma-4-E4B       & \multicolumn{2}{c}{local GPU} & \texttt{google/gemma-4-e4b} \\
\bottomrule
\end{tabular}
\caption{Per-token API pricing (USD/M tokens, May~2026) and model snapshots. The four open-weight models run on $2\times$ NVIDIA L40S 48GB at \texttt{bfloat16}; the full open-weight evaluation across all five benchmarks and both channels took approximately 425 GPU-hours.}
\label{tab:pricing}
\end{table*}

\subsection{Dataset Statistics and Licenses}
\label{app:dataset_stats}

Per-dataset sizes, splits, answer formats, mean L0 token lengths, and licenses are in Table~\ref{tab:dataset_stats}.

\begin{table*}[!htbp]
\centering
\small
\setlength{\tabcolsep}{6pt}
\begin{tabular}{lrlllc}
\toprule
\textbf{Dataset} & $n$ & \textbf{Split} & \textbf{Format} & \textbf{Mean toks} & \textbf{License} \\
\midrule
GSM8K         & 1{,}319 & test & numeric   & 64  & MIT \\
BoolQ         & 3{,}270 & val  & yes/no    & 117 & CC-BY-SA-3.0 \\
ARC-Easy      & 2{,}376 & test & MC (A--D) & 49  & CC-BY-SA-4.0 \\
CommonsenseQA & 1{,}221 & val  & MC (A--E) & 33  & MIT \\
MMLU-STEM     & 3{,}279 & test & MC (A--D) & 85  & MIT \\
\bottomrule
\end{tabular}
\caption{Dataset statistics. Mean tokens at L0 under the GPT-4o tokenizer. All datasets used under their published licenses.}
\label{tab:dataset_stats}
\end{table*}

\subsection{Inference Configuration}
\label{app:inference_config}

All eight models use greedy decoding under identical settings for both conditions; per-level decoder budgets are the only level-dependent parameter (Table~\ref{tab:inference_config}).

\begin{table*}[!htbp]
\centering
\footnotesize
\setlength{\tabcolsep}{4pt}
\begin{tabular}{ll}
\toprule
\textbf{Setting} & \textbf{Value} \\
\midrule
Decoding                       & greedy ($\texttt{temperature}=0$, \texttt{do\_sample=False}) \\
\texttt{max\_new\_tokens} (L0--L4) & $\{400, 300, 200, 150, 20\}$ \\
spaCy                          & 3.8.14 with \texttt{en\_core\_web\_sm} 3.8.0 \\
Open-weight hardware           & 2$\times$ NVIDIA L40S 48GB, \texttt{bfloat16} ($\sim$425 GPU-hours total) \\
NLI judge                      & \texttt{cross-encoder/nli-deberta-v3-base} \\
Embedding model (cosine)       & \texttt{sentence-transformers/all-MiniLM-L6-v2} \citep{reimers-gurevych-2019-sentence} \\
Random seed                    & 42 (where applicable) \\
\bottomrule
\end{tabular}
\caption{Inference configuration. Identical across Conditions A and B.}
\label{tab:inference_config}
\end{table*}

\subsection{Token Accounting}

Mean tokens per level for Claude Haiku~4.5, Condition~A, CommonsenseQA (Table~\ref{tab:token_accounting}). Input tokens fall at L1--L2 but output expansion produces a net increase at L1 and marginal savings at L2; reductions appear at L3--L4, where entailment has already declined.

\begin{table}[ht]
\centering
\footnotesize
\setlength{\tabcolsep}{6pt}
\resizebox{\columnwidth}{!}{%
\begin{tabular}{lrrrrr}
\toprule
\textbf{Level} & \textbf{Mean in} & \textbf{Mean out} &
\textbf{Total} & \textbf{Out/In} & \textbf{Total $\Delta$} \\
\midrule
L0 & 62.8 & 170.3 & 233.1 & 2.71 & baseline \\
L1 & 53.6 & 192.4 & 246.0 & 3.59 & $+$5.5\% \\
L2 & 40.7 & 180.9 & 221.6 & 4.44 & $-$4.9\% \\
L3 & 36.5 & 146.7 & 183.2 & 4.02 & $-$21.4\% \\
L4 & 35.9 &  20.0 &  55.9 & 0.56 & $-$76.0\% \\
\bottomrule
\end{tabular}}
\caption{Token accounting for Claude Haiku~4.5, Condition~A, CommonsenseQA ($n=1{,}221$). Total tokens rise at L1 from output expansion.}
\label{tab:token_accounting}
\end{table}

\subsection{Cost Estimates}

Cost model applied to Claude Haiku~4.5, Condition~A, CommonsenseQA (Table~\ref{tab:cost_estimates}). Cost rises at L1 from output expansion; reductions appear only at higher levels where entailment has fallen.

\begin{table}[ht]
\centering
\footnotesize
\setlength{\tabcolsep}{6pt}
\resizebox{\columnwidth}{!}{%
\begin{tabular}{lrrrrc}
\toprule
\textbf{Lv} & \textbf{In cost} & \textbf{Out cost} &
\textbf{Total cost} & \textbf{Cost $\Delta$} & \textbf{NLI\%} \\
 & \multicolumn{3}{c}{\footnotesize(\$/M items)} & & \\
\midrule
L0 & 62.8 & 851.5 & 914.3 & baseline  & 100.0 \\
L1 & 53.6 & 962.0 & 1015.6 & $+$11.1\% &  50.8 \\
L2 & 40.7 & 904.5 & 945.2 &  $+$3.4\% &  36.4 \\
L3 & 36.5 & 733.5 & 770.0 & $-$15.8\% &  21.9 \\
L4 & 35.9 & 100.0 & 135.9 & $-$85.1\% &  --- \\
\bottomrule
\end{tabular}}
\caption{Estimated inference cost per million items for Claude Haiku~4.5, Condition~A, CommonsenseQA. Lower cost does not imply higher preservation.}
\label{tab:cost_estimates}
\end{table}

Per-model, per-benchmark breakdowns are in Appendix~\ref{app:full}; the input/output token split reproduces from the per-token rates of Table~\ref{tab:pricing}.

\subsection{Open-Weight Cost Projection}
\label{app:open_weight_projection}

The four open-weight models (Qwen2.5-VL-7B, Qwen3.5-9B, DeepSeek-R1-Distill-Qwen-7B, Gemma-4-E4B) have no metered API price, so we project their measured input/output token counts onto a six-tier panel of May~2026 public pricing (DSv4 Flash, DSv4 Pro, Haiku~4.5, GPT-4o, GPT-5.4, Sonnet~4.6) and average across tiers. At L1 Cond~B the mean projected savings are $2.5\times$ on Qwen2.5-VL-7B (the largest open-weight saving in the panel), $1.18\times$ on Qwen3.5-9B, $1.17\times$ on DeepSeek-R1-Distill, and $2.09\times$ on Gemma-4-E4B, all in the same direction as the API panel; the two reasoning-distilled models save less because their unconstrained generations are already short. Full per-tier, per-cell numbers are released in our repository.

\subsection{Decoder-Truncation Check at L1 Condition~B}

Fraction of items whose output hit the L1 \texttt{max\_new\_tokens}=300 ceiling: 0.9\% (GPT-4o), 7.3\% (Haiku~4.5), 0.9\% (Qwen2.5-VL-7B). The L1-B cost savings of Finding~1 reflect natural stopping, not truncation.

\FloatBarrier
\section{Per-Level Results Tables}
\label{app:full}

\begin{figure*}[!htbp]
\centering
\includegraphics[width=\textwidth]{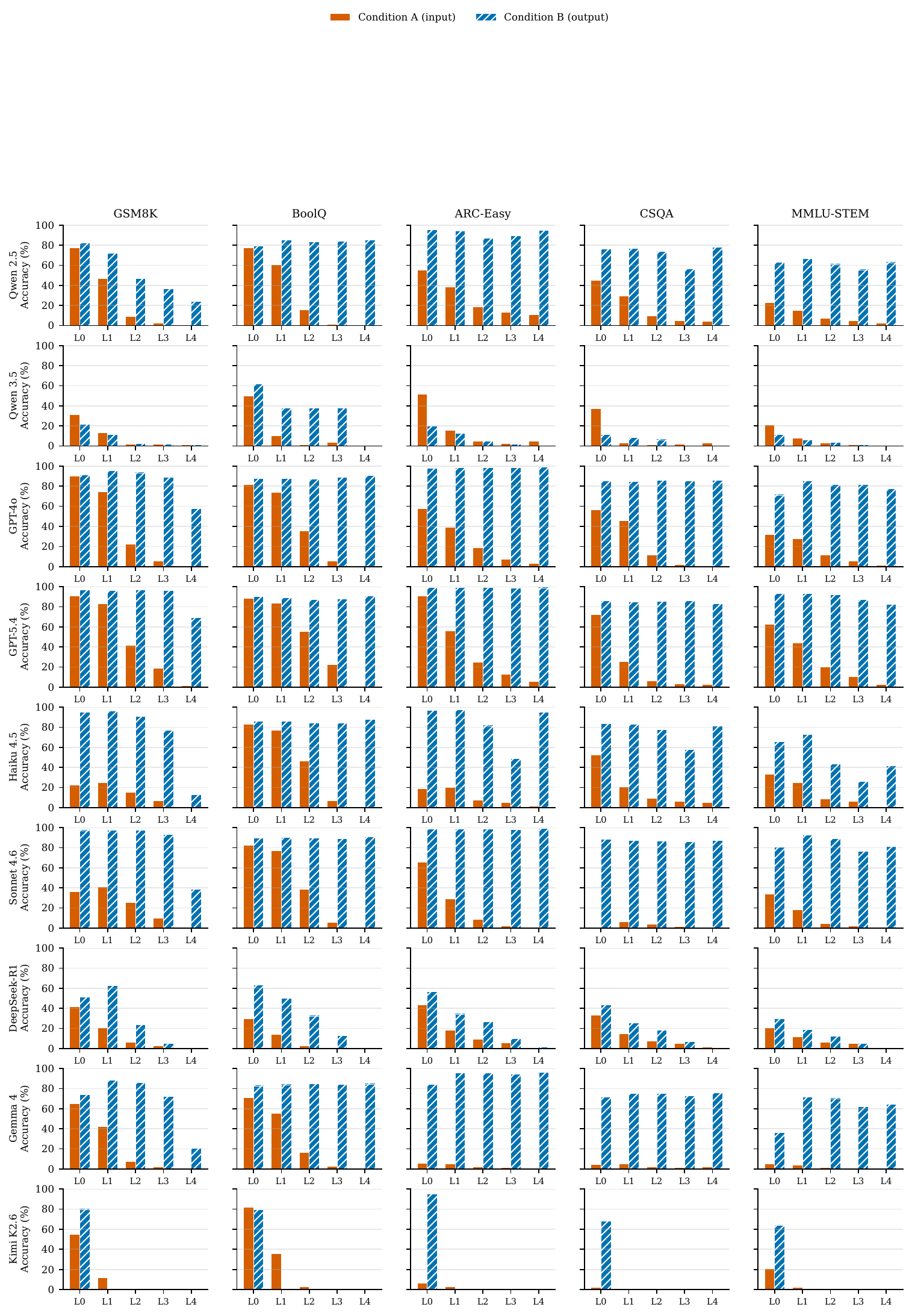}
\caption{Per-level accuracy across the eight evaluated models and five benchmarks under both conditions. Solid: Condition~A. Dashed: Condition~B. The per-level bidirectional NLI grid is released alongside the artifact bundle.}
\label{fig:appendix_accuracy_grid}
\end{figure*}

Per-level accuracy and bidirectional NLI for all eight models (Tables~\ref{tab:acc_summary_A}--\ref{tab:acc_summary_B}). Colored deltas show the change against each model's L0 baseline (red~$\downarrow$ degradation, green~$\uparrow$ improvement). Green Acc deltas with red NLI deltas on classification benchmarks under Condition~B quantify Finding~2; DeepSeek-R1 shows the inverse pattern (\S\ref{sec:finding-dissociation}). Figure~\ref{fig:appendix_accuracy_grid} visualizes the accuracy data as per-model, per-dataset curves.

\begin{table*}[!htbp]
\centering
\footnotesize
\setlength{\tabcolsep}{3pt}
\resizebox{\textwidth}{!}{%
\begin{tabular}{ll cc cc cc cc cc}
\toprule
 &  & \multicolumn{2}{c}{\textbf{GSM8K}} & \multicolumn{2}{c}{\textbf{BoolQ}} & \multicolumn{2}{c}{\textbf{ARC-Easy}} & \multicolumn{2}{c}{\textbf{CSQA}} & \multicolumn{2}{c}{\textbf{MMLU-STEM}} \\
\cmidrule(lr){3-4} \cmidrule(lr){5-6} \cmidrule(lr){7-8} \cmidrule(lr){9-10} \cmidrule(lr){11-12}
\textbf{Model} & \textbf{Lv} & Acc $\uparrow$ & NLI $\uparrow$ & Acc $\uparrow$ & NLI $\uparrow$ & Acc $\uparrow$ & NLI $\uparrow$ & Acc $\uparrow$ & NLI $\uparrow$ & Acc $\uparrow$ & NLI $\uparrow$ \\
\midrule
\multirow{5}{*}{Qwen2.5-VL-7B}
  & L0 & 77.4 & 100.0 & 77.4 & 100.0 & 55.3 & 100.0 & 45.1 & 100.0 & 22.7 & 100.0 \\
  & L1 & 46.8\rdn{30.6} & 71.2\rdn{28.8} & 60.9\rdn{16.5} & 22.8\rdn{77.2} & 38.3\rdn{17.0} & 23.1\rdn{76.9} & 29.7\rdn{15.4} & 25.2\rdn{74.8} & 14.8\rdn{7.9}  & 30.8\rdn{69.2} \\
  & L2 &  8.9\rdn{68.5} & 44.9\rdn{55.1} & 15.9\rdn{61.5} & 10.8\rdn{89.2} & 19.0\rdn{36.3} & 10.7\rdn{89.3} & 10.0\rdn{35.1} & 16.1\rdn{83.9} &  7.2\rdn{15.5} & 16.7\rdn{83.3} \\
  & L3 &  2.4\rdn{75.0} & 27.7\rdn{72.3} &  1.0\rdn{76.4} &  7.2\rdn{92.8} & 13.1\rdn{42.2} &  6.6\rdn{93.4} &  5.2\rdn{39.9} &  7.3\rdn{92.7} &  4.9\rdn{17.8} & 10.2\rdn{89.8} \\
  & L4 &  0.0\rdn{77.4} & ---            &  0.2\rdn{77.2} & ---            & 10.9\rdn{44.4} & ---            &  4.2\rdn{40.9} & ---            &  2.7\rdn{20.0} & ---            \\
\midrule
\multirow{5}{*}{Qwen3.5-9B}
  & L0 & 31.1           & 100.0          & 50.0           & 100.0          & 52.0           & 100.0          & 37.3           & 100.0          & 20.9           & 100.0          \\
  & L1 & 13.3\rdn{17.8} & 48.6\rdn{51.4} & 10.5\rdn{39.5} & 50.9\rdn{49.1} & 16.0\rdn{36.0} & 25.2\rdn{74.8} &  3.2\rdn{34.1} & 23.3\rdn{76.7} &  8.0\rdn{12.9} & 31.0\rdn{69.0} \\
  & L2 &  1.7\rdn{29.4} & 37.5\rdn{62.5} &  1.1\rdn{48.9} & 35.4\rdn{64.6} &  4.7\rdn{47.3} & 15.4\rdn{84.6} &  1.6\rdn{35.7} & 20.9\rdn{79.1} &  3.2\rdn{17.7} & 14.6\rdn{85.4} \\
  & L3 &  1.7\rdn{29.4} & 30.1\rdn{69.9} &  3.9\rdn{46.1} & 31.4\rdn{68.6} &  2.5\rdn{49.5} & 13.2\rdn{86.8} &  1.6\rdn{35.7} & 20.1\rdn{79.9} &  1.5\rdn{19.4} & 11.4\rdn{88.6} \\
  & L4 &  1.1\rdn{30.0} & ---            &  0.0\rdn{50.0} & ---            &  4.8\rdn{47.2} & ---            &  3.3\rdn{34.0} & ---            &  1.0\rdn{19.9} & ---            \\
\midrule
\multirow{5}{*}{GPT-4o}
  & L0 & 89.8 & 100.0 & 81.5 & 100.0 & 58.0 & 100.0 & 56.3 & 100.0 & 32.1 & 100.0 \\
  & L1 & 74.4\rdn{15.4} & 81.0\rdn{19.0} & 74.1\rdn{7.4}  & 35.6\rdn{64.4} & 38.9\rdn{19.1} & 30.1\rdn{69.9} & 45.9\rdn{10.4} & 30.9\rdn{69.1} & 27.7\rdn{4.4}  & 44.9\rdn{55.1} \\
  & L2 & 22.1\rdn{67.7} & 55.0\rdn{45.0} & 35.7\rdn{45.8} & 16.4\rdn{83.6} & 18.9\rdn{39.1} & 15.4\rdn{84.6} & 11.3\rdn{45.0} & 16.1\rdn{83.9} & 11.4\rdn{20.7} & 21.8\rdn{78.2} \\
  & L3 &  5.8\rdn{84.0} & 39.3\rdn{60.7} &  5.7\rdn{75.8} &  7.1\rdn{92.9} &  7.6\rdn{50.4} &  7.7\rdn{92.3} &  2.0\rdn{54.3} &  5.1\rdn{94.9} &  5.8\rdn{26.3} & 12.5\rdn{87.5} \\
  & L4 &  0.4\rdn{89.4} & ---            &  0.0\rdn{81.5} & ---            &  3.2\rdn{54.8} & ---            &  1.0\rdn{55.3} & ---            &  1.6\rdn{30.5} & ---            \\
\midrule
\multirow{5}{*}{GPT-5.4}
  & L0 & 90.8 & 100.0 & 88.5 & 100.0 & 90.5 & 100.0 & 72.3 & 100.0 & 62.7 & 100.0 \\
  & L1 & 82.8\rdn{8.0}  & 68.3\rdn{31.7} & 83.5\rdn{5.0}  & 39.5\rdn{60.5} & 56.1\rdn{34.4} & 34.6\rdn{65.4} & 25.1\rdn{47.2} & 28.9\rdn{71.1} & 44.2\rdn{18.6} & 50.4\rdn{49.6} \\
  & L2 & 41.5\rdn{49.4} & 40.2\rdn{59.8} & 55.3\rdn{33.1} & 23.1\rdn{76.9} & 24.5\rdn{66.0} & 21.0\rdn{79.0} &  6.0\rdn{66.3} & 16.8\rdn{83.2} & 20.0\rdn{42.7} & 29.6\rdn{70.4} \\
  & L3 & 18.9\rdn{71.9} & 24.6\rdn{75.4} & 22.1\rdn{66.3} & 18.1\rdn{81.9} & 12.7\rdn{77.8} & 15.2\rdn{84.8} &  3.4\rdn{69.0} &  9.8\rdn{90.2} & 10.6\rdn{52.2} & 23.3\rdn{76.7} \\
  & L4 &  1.1\rdn{89.7} & ---            &  0.7\rdn{87.8} & ---            &  5.5\rdn{85.0} & ---            &  2.9\rdn{69.5} & ---            &  2.7\rdn{60.1} & ---            \\
\midrule
\multirow{5}{*}{Haiku 4.5}
  & L0 & 22.4 & 100.0 & 82.9 & 100.0 & 19.1 & 100.0 & 52.3 & 100.0 & 33.1 & 100.0 \\
  & L1 & 24.6\rup{2.2}  & 62.8\rdn{37.2} & 77.2\rdn{5.7}  & 37.2\rdn{62.8} & 20.2\rup{1.1}  & 46.3\rdn{53.7} & 20.4\rdn{31.9} & 50.8\rdn{49.2} & 25.0\rdn{8.1}  & 44.0\rdn{56.0} \\
  & L2 & 14.9\rdn{7.5}  & 29.7\rdn{70.3} & 46.4\rdn{36.5} & 20.8\rdn{79.2} &  7.3\rdn{11.8} & 29.4\rdn{70.6} &  9.2\rdn{43.1} & 36.4\rdn{63.6} &  8.7\rdn{24.4} & 25.7\rdn{74.3} \\
  & L3 &  6.8\rdn{15.6} & 19.0\rdn{81.0} &  6.7\rdn{76.2} & 18.9\rdn{81.1} &  4.8\rdn{14.3} & 19.7\rdn{80.3} &  6.3\rdn{46.0} & 21.9\rdn{78.1} &  6.0\rdn{27.1} & 19.8\rdn{80.2} \\
  & L4 &  0.1\rdn{22.3} & ---            &  0.0\rdn{82.9} & ---            &  1.6\rdn{17.5} & ---            &  4.8\rdn{47.5} & ---            &  1.0\rdn{32.1} & ---            \\
\midrule
\multirow{5}{*}{Sonnet 4.6}
  & L0 & 36.2           & 100.0          & 82.6           & 100.0          & 65.6           & 100.0          &  1.0$^{\dagger}$           & 100.0          & 34.1           & 100.0          \\
  & L1 & 41.2\rup{5.0}  & 61.0\rdn{39.0} & 77.0\rdn{5.6}  & 44.5\rdn{55.5} & 29.2\rdn{36.4} & 43.5\rdn{56.5} &  6.0\rup{5.0}  & 30.0\rdn{70.0} & 18.5\rdn{15.6} & 44.1\rdn{55.9} \\
  & L2 & 25.5\rdn{10.6} & 36.2\rdn{63.8} & 38.7\rdn{43.9} & 24.6\rdn{75.4} &  8.8\rdn{56.7} & 26.3\rdn{73.7} &  3.9\rup{2.9}  & 16.9\rdn{83.1} &  4.6\rdn{29.5} & 29.4\rdn{70.6} \\
  & L3 &  9.9\rdn{26.2} & 27.0\rdn{73.0} &  5.5\rdn{77.1} & 19.1\rdn{80.9} &  1.9\rdn{63.7} & 18.6\rdn{81.4} &  1.7\rup{0.7}  & 12.1\rdn{87.9} &  2.1\rdn{31.9} & 22.6\rdn{77.4} \\
  & L4 &  0.4\rdn{35.8} & ---            &  0.0\rdn{82.5} & ---            &  1.1\rdn{64.5} & ---            &  0.7\rdn{0.3}  & ---            &  0.6\rdn{33.5} & ---            \\
\midrule
\multirow{5}{*}{DeepSeek-R1}
  & L0 & 41.9 & 100.0 & 30.0 & 100.0 & 43.4 & 100.0 & 33.3 & 100.0 & 20.9 & 100.0 \\
  & L1 & 20.7\rdn{21.2} & 79.5\rdn{20.5} & 14.4\rdn{15.6} & 12.7\rdn{87.3} & 18.4\rdn{25.1} & 18.8\rdn{81.2} & 14.8\rdn{18.5} & 35.4\rdn{64.6} & 11.5\rdn{9.4}  & 23.2\rdn{76.8} \\
  & L2 &  6.2\rdn{35.7} & 54.9\rdn{45.1} &  2.7\rdn{27.3} &  9.6\rdn{90.4} &  9.2\rdn{34.3} & 13.8\rdn{86.2} &  7.7\rdn{25.6} & 33.5\rdn{66.5} &  6.3\rdn{14.6} & 12.9\rdn{87.1} \\
  & L3 &  2.5\rdn{39.4} & 48.1\rdn{51.9} &  0.5\rdn{29.5} & 11.3\rdn{88.7} &  6.0\rdn{37.5} & 14.1\rdn{85.9} &  4.9\rdn{28.4} & 32.6\rdn{67.4} &  4.8\rdn{16.1} & 10.9\rdn{89.1} \\
  & L4 &  0.2\rdn{41.7} & ---            &  0.0\rdn{30.0} & ---            &  1.1\rdn{42.3} & ---            &  1.6\rdn{31.7} & ---            &  0.7\rdn{20.2} & ---            \\
\midrule
\multirow{5}{*}{Gemma-4-E4B}
  & L0 & 65.1 & 100.0 & 71.2 & 100.0 & 6.0$^{\dagger}$ & 100.0 & 4.4$^{\dagger}$ & 100.0 & 5.4$^{\dagger}$ & 100.0 \\
  & L1 & 42.2\rdn{22.9} & 67.0\rdn{33.0} & 55.6\rdn{15.6} & 38.0\rdn{62.0} & 5.4\rdn{0.6} & 33.3\rdn{66.7} & 5.1\rup{0.7} & 47.2\rdn{52.8} & 4.0\rdn{1.4} & 37.1\rdn{62.9} \\
  & L2 & 7.7\rdn{57.4} & 26.4\rdn{73.6} & 16.8\rdn{54.4} & 15.4\rdn{84.6} & 2.4\rdn{3.7} & 17.7\rdn{82.3} & 2.1\rdn{2.3} & 29.4\rdn{70.6} & 1.6\rdn{3.8} & 14.5\rdn{85.5} \\
  & L3 & 2.2\rdn{62.9} & 12.6\rdn{87.4} & 2.5\rdn{68.7} & 8.2\rdn{91.8} & 1.6\rdn{4.5} & 9.3\rdn{90.7} & 1.2\rdn{3.2} & 17.4\rdn{82.6} & 1.2\rdn{4.1} & 8.0\rdn{92.0} \\
  & L4 & 0.1\rdn{65.0} & --- & 0.1\rdn{71.1} & --- & 1.0\rdn{5.0} & --- & 2.1\rdn{2.3} & --- & 0.5\rdn{4.9} & --- \\
\bottomrule
\end{tabular}}
\caption{Per-level task accuracy (\%) and bidirectional NLI entailment rate (\%) under Condition~A (input compression) for all eight evaluated models. NLI~\% is the rate at which a generation entails its L0 counterpart bidirectionally; L0 is 100~\% by construction. Colored subscripts show the change relative to each model's L0 baseline (red~$\downarrow$~for degradation, green~$\uparrow$~for improvement). L4 NLI not reported; see~\S\ref{sec:metrics}. \\ \footnotesize{$^{\dagger}$ L0 extraction rate $< 0.95$; affected cells are audited in Appendix~G.}}
\label{tab:acc_summary_A}
\end{table*}

\begin{table*}[!htbp]
\centering
\footnotesize
\setlength{\tabcolsep}{3pt}
\resizebox{\textwidth}{!}{%
\begin{tabular}{ll cc cc cc cc cc}
\toprule
 &  & \multicolumn{2}{c}{\textbf{GSM8K}} & \multicolumn{2}{c}{\textbf{BoolQ}} & \multicolumn{2}{c}{\textbf{ARC-Easy}} & \multicolumn{2}{c}{\textbf{CSQA}} & \multicolumn{2}{c}{\textbf{MMLU-STEM}} \\
\cmidrule(lr){3-4} \cmidrule(lr){5-6} \cmidrule(lr){7-8} \cmidrule(lr){9-10} \cmidrule(lr){11-12}
\textbf{Model} & \textbf{Lv} & Acc $\uparrow$ & NLI $\uparrow$ & Acc $\uparrow$ & NLI $\uparrow$ & Acc $\uparrow$ & NLI $\uparrow$ & Acc $\uparrow$ & NLI $\uparrow$ & Acc $\uparrow$ & NLI $\uparrow$ \\
\midrule
\multirow{5}{*}{Qwen2.5-VL-7B}
  & L0 & 82.4 & 100.0 & 79.4 & 100.0 & 95.4 & 100.0 & 76.5 & 100.0 & 62.9$^{\dagger}$ & 100.0 \\
  & L1 & 72.3\rdn{10.1} & 64.5\rdn{35.5} & 85.1\rup{5.7}  & 24.6\rdn{75.4} & 94.2\rdn{1.2} & 38.4\rdn{61.6} & 76.7\rup{0.2}  & 44.4\rdn{55.6} & 66.6\rup{3.7}  & 28.5\rdn{71.5} \\
  & L2 & 46.9\rdn{35.5} & 44.4\rdn{55.6} & 83.2\rup{3.8}  & 21.8\rdn{78.2} & 86.8\rdn{8.6} & 36.3\rdn{63.7} & 74.0\rdn{2.5}  & 42.5\rdn{57.5} & 61.0\rdn{1.9}  & 24.5\rdn{75.5} \\
  & L3 & 36.7\rdn{45.7} & 34.1\rdn{65.9} & 84.2\rup{4.8}  & 27.8\rdn{72.2} & 89.2\rdn{6.2} & 23.7\rdn{76.3} & 56.5\rdn{20.0} & 24.3\rdn{75.7} & 55.5\rdn{7.4}  & 16.7\rdn{83.3} \\
  & L4 & 23.8\rdn{58.6} & ---            & 85.0\rup{5.6}  & ---            & 94.8\rdn{0.6} & ---            & 78.2\rup{1.7}  & ---            & 62.7\rdn{0.2}  & ---            \\
\midrule
\multirow{5}{*}{Qwen3.5-9B}
  & L0 & 21.5           & 100.0          & 62.0           & 100.0          & 19.9           & 100.0          & 11.5           & 100.0          & 11.7           & 100.0          \\
  & L1 & 11.4\rdn{10.1} & 66.9\rdn{33.1} & 37.8\rdn{24.2} & 40.9\rdn{59.1} & 12.8\rdn{7.1}  & 21.4\rdn{78.6} &  8.7\rdn{2.8}  & 34.7\rdn{65.3} &  5.9\rdn{5.8}  & 20.3\rdn{79.7} \\
  & L2 &  2.7\rdn{18.8} & 16.4\rdn{83.6} & 37.8\rdn{24.2} &  7.3\rdn{92.7} &  5.0\rdn{14.9} &  5.9\rdn{94.1} &  6.4\rdn{5.1}  &  9.1\rdn{90.9} &  3.5\rdn{8.2}  &  5.3\rdn{94.7} \\
  & L3 &  2.0\rdn{19.5} & 17.4\rdn{82.6} & 37.8\rdn{24.2} & 14.3\rdn{85.7} &  1.8\rdn{18.1} &  6.4\rdn{93.6} &  0.7\rdn{10.8} & 15.9\rdn{84.1} &  1.2\rdn{10.5} &  6.6\rdn{93.4} \\
  & L4 &  1.1\rdn{20.4} & ---            &  0.0\rdn{62.0} & ---            &  0.0\rdn{19.9} & ---            &  0.0\rdn{11.5} & ---            &  0.0\rdn{11.7} & ---            \\
\midrule
\multirow{5}{*}{GPT-4o}
  & L0 & 91.5 & 100.0 & 87.6 & 100.0 & 97.9 & 100.0 & 85.3 & 100.0 & 71.2$^{\dagger}$ & 100.0 \\
  & L1 & 95.6\rup{4.1}  & 80.1\rdn{19.9} & 87.8\rup{0.2} & 36.2\rdn{63.8} & 98.8\rup{0.9} & 37.2\rdn{62.8} & 84.7\rdn{0.6} & 57.4\rdn{42.6} & 85.1\rup{13.9} & 30.7\rdn{69.3} \\
  & L2 & 93.4\rup{1.9}  & 80.5\rdn{19.5} & 87.2\rdn{0.4} & 40.5\rdn{59.5} & 98.5\rup{0.6} & 34.9\rdn{65.1} & 85.9\rup{0.6} & 57.5\rdn{42.5} & 81.9\rup{10.7} & 32.5\rdn{67.5} \\
  & L3 & 88.9\rdn{2.6}  & 78.8\rdn{21.2} & 89.1\rup{1.5} & 34.4\rdn{65.6} & 98.7\rup{0.8} & 35.1\rdn{64.9} & 85.2\rdn{0.1} & 55.4\rdn{44.6} & 81.5\rup{10.3} & 31.5\rdn{68.5} \\
  & L4 & 57.5\rdn{34.0} & ---            & 90.8\rup{3.2} & ---            & 98.9\rup{1.0} & ---            & 85.8\rup{0.5} & ---            & 77.4\rup{6.2}  & ---            \\
\midrule
\multirow{5}{*}{GPT-5.4}
  & L0 & 96.9 & 100.0 & 90.1 & 100.0 & 99.1 & 100.0 & 85.8 & 100.0 & 93.0 & 100.0 \\
  & L1 & 96.3\rdn{0.6}  & 65.4\rdn{34.6} & 88.9\rdn{1.2} & 35.8\rdn{64.2} & 99.1          & 30.6\rdn{69.4} & 84.8\rdn{1.0} & 40.6\rdn{59.4} & 93.2\rup{0.2}  & 46.3\rdn{53.7} \\
  & L2 & 96.7\rdn{0.2}  & 65.7\rdn{34.3} & 87.3\rdn{2.8} & 46.7\rdn{53.3} & 99.2\rup{0.1} & 35.0\rdn{65.0} & 85.4\rdn{0.4} & 41.9\rdn{58.1} & 91.8\rdn{1.3}  & 47.2\rdn{52.8} \\
  & L3 & 95.9\rdn{1.0}  & 66.6\rdn{33.4} & 87.9\rdn{2.2} & 36.5\rdn{63.5} & 98.8\rdn{0.3} & 33.4\rdn{66.6} & 85.8          & 42.0\rdn{58.0} & 86.9\rdn{6.1}  & 46.4\rdn{53.6} \\
  & L4 & 69.2\rdn{27.7} & ---            & 90.7\rup{0.6} & ---            & 98.8\rdn{0.3} & ---            & 83.2\rdn{2.6} & ---            & 82.1\rdn{10.9} & ---            \\
\midrule
\multirow{5}{*}{Haiku 4.5}
  & L0 & 95.1 & 100.0 & 85.7 & 100.0 & 96.9 & 100.0 & 83.7 & 100.0 & 65.8 & 100.0 \\
  & L1 & 95.9\rup{0.8}  & 58.5\rdn{41.5} & 86.2\rup{0.5} & 49.4\rdn{50.6} & 97.1\rup{0.2}  & 36.9\rdn{63.1} & 82.7\rdn{1.0}  & 67.1\rdn{32.9} & 72.6\rup{6.8}  & 32.9\rdn{67.1} \\
  & L2 & 90.8\rdn{4.3}  & 58.4\rdn{41.6} & 84.2\rdn{1.5} & 52.8\rdn{47.2} & 81.5\rdn{15.4} & 35.0\rdn{65.0} & 77.5\rdn{6.2}  & 69.7\rdn{30.3} & 43.5\rdn{22.3} & 29.0\rdn{71.0} \\
  & L3 & 76.7\rdn{18.4} & 52.0\rdn{48.0} & 83.9\rdn{1.8} & 51.4\rdn{48.6} & 48.9\rdn{48.0} & 28.7\rdn{71.3} & 57.9\rdn{25.8} & 65.1\rdn{34.9} & 25.7\rdn{40.1} & 21.8\rdn{78.2} \\
  & L4 & 12.7\rdn{82.4} & ---            & 87.5\rup{1.8} & ---            & 95.2\rdn{1.7}  & ---            & 81.3\rdn{2.4}  & ---            & 41.7\rdn{24.1} & ---            \\
\midrule
\multirow{5}{*}{Sonnet 4.6}
  & L0 & 97.1           & 100.0          & 89.7           & 100.0          & 98.7           & 100.0          & 88.2           & 100.0          & 80.4$^{\dagger}$           & 100.0          \\
  & L1 & 97.4\rup{0.3}  & 59.3\rdn{40.7} & 90.0\rup{0.3} & 32.7\rdn{67.3} & 98.7           & 20.7\rdn{79.3} & 87.3\rdn{0.9}  & 42.2\rdn{57.8} & 92.7\rup{12.3} & 33.5\rdn{66.5} \\
  & L2 & 97.3\rup{0.2}  & 57.8\rdn{42.2} & 89.4\rdn{0.3} & 37.6\rdn{62.4} & 98.9\rup{0.2} & 23.5\rdn{76.5} & 86.8\rdn{1.4}  & 48.9\rdn{51.1} & 89.0\rup{8.6}  & 33.9\rdn{66.1} \\
  & L3 & 93.0\rdn{4.1}  & 59.6\rdn{40.4} & 89.0\rdn{0.7} & 36.2\rdn{63.8} & 98.2\rdn{0.5} & 29.3\rdn{70.7} & 85.8\rdn{2.4}  & 53.1\rdn{46.9} & 76.5\rdn{3.9}  & 34.2\rdn{65.8} \\
  & L4 & 38.5\rdn{58.6} & ---            & 90.9\rup{1.2} & ---            & 99.2\rup{0.5} & ---            & 87.4\rdn{0.8}  & ---            & 81.2\rup{0.8}  & ---            \\
\midrule
\multirow{5}{*}{DeepSeek-R1}
  & L0 & 51.3 & 100.0 & 63.4 & 100.0 & 56.7 & 100.0 & 43.6 & 100.0 & 29.9 & 100.0 \\
  & L1 & 62.5\rup{11.2} & 74.4\rdn{25.6} & 50.0\rdn{13.4} & 38.4\rdn{61.6} & 34.2\rdn{22.5} & 29.4\rdn{70.6} & 25.6\rdn{18.0} & 47.2\rdn{52.8} & 19.1\rdn{10.8} & 27.9\rdn{72.1} \\
  & L2 & 23.8\rdn{27.4} & 69.0\rdn{31.0} & 32.4\rdn{31.0} & 38.1\rdn{61.9} & 26.5\rdn{30.2} & 31.1\rdn{68.9} & 18.3\rdn{25.3} & 51.1\rdn{48.9} & 12.3\rdn{17.6} & 28.2\rdn{71.8} \\
  & L3 &  5.0\rdn{46.2} & 64.3\rdn{35.7} & 13.0\rdn{50.4} & 37.0\rdn{63.0} &  9.8\rdn{46.9} & 29.5\rdn{70.5} &  7.1\rdn{36.4} & 52.7\rdn{47.3} &  5.1\rdn{24.8} & 25.0\rdn{75.0} \\
  & L4 &  0.3\rdn{50.9} & ---            &  0.1\rdn{63.3} & ---            &  0.6\rdn{56.1} & ---            &  0.7\rdn{42.9} & ---            &  0.4\rdn{29.5} & ---            \\
\midrule
\multirow{5}{*}{Gemma-4-E4B}
  & L0 & 74.0 & 100.0 & 82.8 & 100.0 & 84.3 & 100.0 & 71.7 & 100.0 & 36.5$^{\dagger}$ & 100.0 \\
  & L1 & 88.6\rup{14.6} & 54.2\rdn{45.8} & 84.0\rup{1.2} & 46.4\rdn{53.6} & 95.8\rup{11.5} & 28.6\rdn{71.4} & 75.1\rup{3.4} & 54.7\rdn{45.3} & 71.9\rup{35.3} & 19.9\rdn{80.1} \\
  & L2 & 86.3\rup{12.3} & 51.9\rdn{48.1} & 84.9\rup{2.1} & 43.9\rdn{56.1} & 95.7\rup{11.4} & 25.4\rdn{74.6} & 75.4\rup{3.8} & 50.4\rdn{49.6} & 70.2\rup{33.7} & 15.5\rdn{84.5} \\
  & L3 & 72.1\rdn{1.9} & 47.7\rdn{52.3} & 84.0\rup{1.2} & 39.8\rdn{60.2} & 94.3\rup{10.0} & 28.5\rdn{71.5} & 72.8\rup{1.1} & 51.4\rdn{48.6} & 62.1\rup{25.6} & 16.8\rdn{83.2} \\
  & L4 & 20.6\rdn{53.4} & --- & 84.6\rup{1.8} & --- & 96.2\rup{11.9} & --- & 75.8\rup{4.1} & --- & 64.6\rup{28.1} & --- \\
\bottomrule
\end{tabular}}
\caption{Per-level task accuracy (\%) and bidirectional NLI entailment rate (\%) under Condition~B (output constraint) for all eight evaluated models. NLI~\% is the rate at which a generation entails its L0 counterpart bidirectionally; L0 is 100~\% by construction. Colored subscripts show the change relative to each model's L0 baseline (red~$\downarrow$~for degradation, green~$\uparrow$~for improvement). The contrast between green Acc deltas and red NLI deltas on classification benchmarks under non-reasoning models is the surface-text divergence of Finding~2. L4 NLI not reported; see~\S\ref{sec:metrics}. \\ \footnotesize{$^{\dagger}$ L0 extraction rate $< 0.95$; affected cells are audited in Appendix~G.}}
\label{tab:acc_summary_B}
\end{table*}

\subsection{Threshold and Task-Type Comparisons}
\label{app:threshold}

\begin{figure*}[!htbp]
\centering
\includegraphics[width=\textwidth]{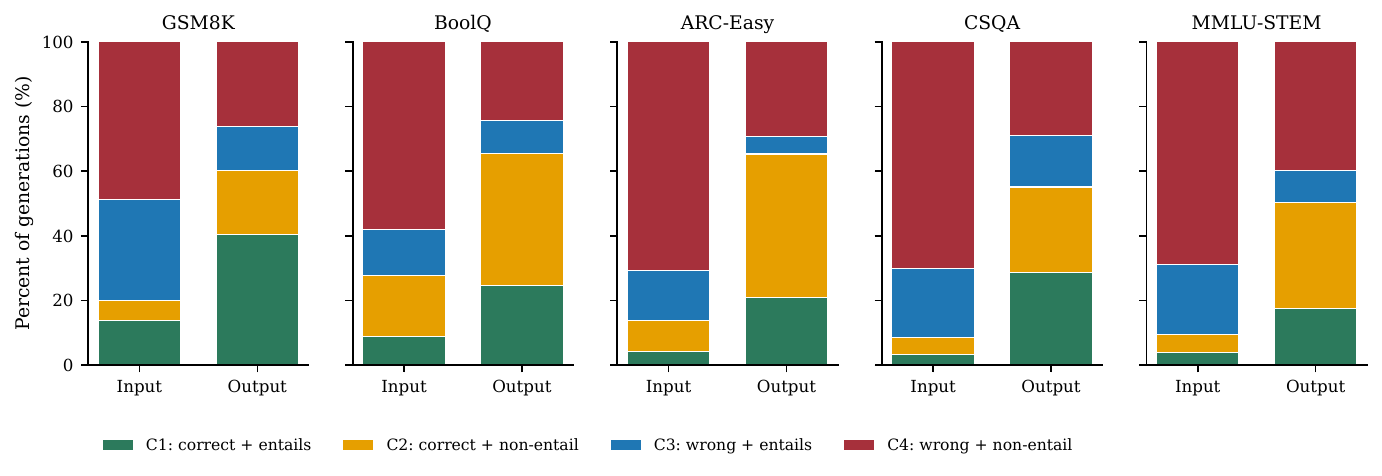}
\caption{$2\times2$ dissociation by dataset, aggregated across L1--L3 and the eight evaluated models. Each panel shows Condition~A and Condition~B side by side; bar segments are the $C_1$--$C_4$ outcome shares.}
\label{fig:appendix_dissociation_by_dataset}
\end{figure*}

Smallest level $L_c$ at which accuracy ($L_c^{\text{acc}}$) and NLI ($L_c^{\text{sem}}$) cross the degradation criterion (Table~\ref{tab:lc_comparison}): we define degradation as accuracy falling at least $5$~pp below the L0 baseline ($L_c^{\text{acc}}$, scanned over L1--L4) and bidirectional NLI falling at least $15$~pp below the L0 anchor ($L_c^{\text{sem}}$, scanned over L1--L3 since L4 carries no NLI score); ``---'' marks a cell where the threshold is never crossed in that range (non-evaluable for the ordering). Of 80 `(model, dataset, condition)' cells, 60 are evaluable (both thresholds crossed in range), and on the 8-model panel under strict bidirectional NLI, $L_c^{\text{sem}} \leq L_c^{\text{acc}}$ on all 60. Forward-only NLI violates the ordering on 2 of 60; cosine on 36 of 60, confirming cosine is non-monotone. The rows in Table~\ref{tab:lc_comparison} are a representative three-model excerpt; the 60-of-60 figure is computed over the full eight-model panel (released data). Table~\ref{tab:task_comparison} aggregates L1 non-preservation by benchmark; Figure~\ref{fig:appendix_dissociation_by_dataset} breaks the $2\times2$ dissociation down per-benchmark.

\begin{table*}[!htbp]
\centering
\small
\setlength{\tabcolsep}{6pt}
\begin{tabular}{llcccc}
\toprule
\textbf{Benchmark} & \textbf{Model} &
\multicolumn{2}{c}{\textbf{Cond A}} &
\multicolumn{2}{c}{\textbf{Cond B}} \\
 & & $L_c^{\text{acc}}$ & $L_c^{\text{sem}}$ &
     $L_c^{\text{acc}}$ & $L_c^{\text{sem}}$ \\
\midrule
GSM8K        & Qwen2.5-VL-7B  & L1 & L1 & L1  & L1 \\
             & GPT-4o    & L1 & L1 & L4  & L1 \\
             & Haiku 4.5 & L2 & L1 & L3  & L1 \\
BoolQ        & Qwen2.5-VL-7B  & L1 & L1 & --- & L1 \\
             & GPT-4o    & L1 & L1 & --- & L1 \\
             & Haiku 4.5 & L1 & L1 & --- & L1 \\
ARC-Easy     & Qwen2.5-VL-7B  & L1 & L1 & L2  & L1 \\
             & GPT-4o    & L1 & L1 & --- & L1 \\
             & Haiku 4.5 & L2 & L1 & L2  & L1 \\
CommonsenseQA& Qwen2.5-VL-7B  & L1 & L1 & L3  & L1 \\
             & GPT-4o    & L1 & L1 & --- & L1 \\
             & Haiku 4.5 & L1 & L1 & L2  & L1 \\
MMLU-STEM    & Qwen2.5-VL-7B  & L1 & L1 & L3  & L1 \\
             & GPT-4o    & L2 & L1 & --- & L1 \\
             & Haiku 4.5 & L1 & L1 & L2  & L1 \\
\bottomrule
\end{tabular}
\caption{Threshold levels at which accuracy ($L_c^{\text{acc}}$, $\geq 5$~pp below L0, scanned L1--L4) and NLI ($L_c^{\text{sem}}$, $\geq 15$~pp below L0, scanned L1--L3 as L4 has no NLI) first cross the degradation criterion; ``---'' marks a cell where the threshold is never crossed in that range. Of 80 `(model, dataset, condition)' cells, 60 are evaluable (both thresholds crossed in range), and $L_c^{\text{sem}} \leq L_c^{\text{acc}}$ on all 60 on the full eight-model panel under the strict bidirectional NLI criterion; the rows shown are a representative three-model excerpt. The frequent ``---'' under Cond~B reflects accuracy that never degrades by $5$~pp under output compression (Finding~2).}
\label{tab:lc_comparison}
\end{table*}

\begin{table}[ht]
\centering
\small
\begin{tabular}{lcc}
\toprule
\textbf{Benchmark} & \textbf{Cond A} & \textbf{Cond B} \\
\midrule
GSM8K         & 32.6\% & 34.6\% \\
BoolQ         & 64.8\% & 61.9\% \\
ARC-Easy      & 68.1\% & 69.6\% \\
CommonsenseQA & 66.0\% & 51.5\% \\
MMLU-STEM     & 61.8\% & 70.0\% \\
\bottomrule
\end{tabular}
\caption{Mean L1 semantic non-preservation (\%) by benchmark under the headline bidirectional NLI criterion, averaged across the eight evaluated models. The same task-type ordering holds under the alternative criteria of \S\ref{sec:metrics} (Appendix~\ref{app:semantic_robustness}).}
\label{tab:task_comparison}
\end{table}

\FloatBarrier
\section{Judge Reliability and Semantic Robustness}
\label{app:semantic_robustness}

Calibration and robustness evidence for Finding~2: judge reliability by compression level (\S\ref{app:judge_reliability}, Table~\ref{tab:judge_reliability}) and cross-metric replication under eleven additional measures (Table~\ref{tab:alt_metrics}). The headline NLI rate sits near the conservative end of the family.

\subsection{NLI Judge Reliability by Compression Level}
\label{app:judge_reliability}

Calibration uses 70 POS-filtered synthetic positive pairs per level: pairs are semantically equivalent by construction, so non-entailment is attributable to the judge. The Disagree-A rate (NLI fail with cosine $> 0.85$) at L1 Cond~B is 17.1\%, leaving judge-failure cases a clear minority. L1 is judge-reliable (FN $2.9\%$), L2 supplementary (FN $28.6\%$), L3 descriptive only (FN $50.0\%$); L4 is excluded.

\begin{table*}[!htbp]
\centering
\small
\begin{tabular}{lccccc}
\toprule
\textbf{Level} & \textbf{$C_2$ strict (grand)} & \textbf{$C_2$ fwd-only} & \textbf{Disagree-A (Cond B)} & \textbf{Judge FN (calib)} & \textbf{Tag} \\
\midrule
L1 & 42.7\,\% & 38.8\,\% & 17.1\,\% & 2.9\,\%  & Robust (Cond B) \\
L2 & 39.5\,\% & 34.8\,\% & 14.1\,\% & 28.6\,\% & Uncertain \\
L3 & 36.7\,\% & 30.2\,\% & 8.5\,\%  & 50.0\,\% & Unreliable \\
L4 & \multicolumn{5}{c}{excluded from semantic evaluation} \\
\bottomrule
\end{tabular}
\caption{NLI judge reliability by compression level (8-model grand aggregate, Cond~B). \emph{$C_2$ strict}: bidirectional entailment. \emph{$C_2$ fwd-only}: forward direction only. \emph{Disagree-A}: NLI fails but cosine $> 0.85$ (Cond~B). \emph{Judge FN}: false-negative rate on POS-filtered synthetic positive pairs. L1's $2.9\%$ FN underpins the L1 headline.}
\label{tab:judge_reliability}
\end{table*}

\subsection{Robustness Across Alternative Semantic Measures}

We re-scored every (compressed, L0) pair under the eleven complementary $C_2$ measures that, with the headline, make up the twelve: forward-only and soft NLI, three independent NLI judges (BART-large-MNLI, multilingual XNLI, DeBERTa-large), a faithfulness checker (MiniCheck), learned and surface similarity (BLEURT, ROUGE-L, METEOR), an STS cross-encoder, and a QA-based propositional check. Continuous-similarity scores (BERTScore, two sentence-embedding cosines) and an answer-anchored NLI variant are reported alongside for context. Table~\ref{tab:alt_metrics} reports L1 Cond~B aggregates on the 6-non-reasoning panel. All twelve $C_2$ measures report a substantial divergence on the 6-non-reasoning panel (roughly 41--88\%), with the headline bidirectional NLI near the conservative end of the family. Per-metric details and the full per-cell breakdown are released in our repository.

\begin{table*}[!htbp]
\centering
\small
\setlength{\tabcolsep}{4pt}
\begin{tabular}{lll}
\toprule
\textbf{Measure} & \textbf{L1 Cond B value} & \textbf{Type} \\
\midrule
\multicolumn{3}{l}{\emph{The twelve $C_2$ measures (thresholded)}} \\
NLI bidirectional & 51.9\,\% $C_2$ & Thresholded rate \\
NLI forward-only                    & 47.4\,\% $C_2$ & Thresholded rate \\
NLI soft (mean prob $> 0.5$)        & 41.2\,\% $C_2$ & Thresholded rate \\
BART-large-MNLI bidirectional       & 56.1\,\% $C_2$ & Independent NLI judge \\
mDeBERTa-v3-XNLI bidirectional      & 50.2\,\% $C_2$ & Multilingual NLI judge \\
MiniCheck                           & 54.0\,\% $C_2$ & Faithfulness checker \\
DeBERTa-large bidirectional         & 49.5\,\% $C_2$ & Larger NLI variant \\
BLEURT (\texttt{Elron/bleurt-base-128}) & 87.6\,\% $C_2$ (BLEURT $< 0$)  & Learned similarity \\
METEOR                                  & 81.9\,\% $C_2$ (METEOR $< 0.3$) & Paraphrase-aware surface \\
ROUGE-L                                 & 55.9\,\% $C_2$ (ROUGE-L $< 0.3$) & Surface overlap \\
STS cross-encoder                        & 87.2\,\% $C_2$ (STS-pass $< 0.5$) & STS \\
QA-based propositional score         & 60.5\,\% $C_2$ & Content recovery \\
\midrule
\multicolumn{3}{l}{\emph{Reported for context (not counted)}} \\
BERTScore (\texttt{roberta-large})       & 0.858 mean F1 & Token-level \\
Cosine (MiniLM, paper)                   & 0.725 mean cosine & Embedding \\
Cosine (\texttt{intfloat/e5-base-v2})    & 0.904 mean cosine & Embedding, 2nd arch. \\
Answer-anchored NLI                  & 11.4\,\% $C_2$ & Ground truth \\
\midrule
Judge calibration (FN on L1 positives) & 2.9\,\% & Lower bound \\
\bottomrule
\end{tabular}
\caption{Cross-metric agreement at L1 Condition~B (6-non-reasoning panel). The twelve $C_2$ measures all report a substantial divergence on the 6-non-reasoning panel ($41\%$--$88\%$); the headline bidirectional NLI ($51.9\%$ on the 6-non-reasoning panel) sits near the conservative end of the family, which ranges from DeBERTa-large ($49.5\%$ on the 6-non-reasoning panel) to BLEURT ($87.6\%$ on the 6-non-reasoning panel). Continuous-similarity scores (BERTScore, cosine) and the answer-anchored check are shown for context. Answer-anchored NLI is lower because a correct answer usually entails the ground-truth-answer hypothesis even when the surrounding reasoning text diverges.}
\label{tab:alt_metrics}
\end{table*}

\subsection{Length-Controlled NLI Re-Scoring}
\label{app:noise_floor}

Bounds the length-and-register confound. For each (L0, L1-B) pair we truncate L0 to the L1-B wordpiece-token length and re-score with the same judge under the same bidirectional criterion.

\paragraph{Procedure.}
For each (model, dataset, item) tuple at L1-B, tokenize L0, truncate to $\min(|\text{L0}|, |\text{L1-B}|)$, detokenize, re-score in both directions. Same denominator as the Finding~2 headline.

\paragraph{Headline.}
The $C_2$ rate \emph{rises} by $+28.4$~pp on the 6-non-reasoning panel and $+21.6$~pp on the 8-model aggregate under length-matched scoring (Table~\ref{tab:length_controlled}). Reasoning models (DeepSeek-R1, Qwen3.5-9B) move by $\le 7$~pp because their L0 outputs are already short.

\begin{table}[ht]
\centering
\small
\setlength{\tabcolsep}{6pt}
\resizebox{\columnwidth}{!}{%
\begin{tabular}{lccc}
\toprule
\textbf{Panel} & \textbf{Original $C_2$} & \textbf{Length-controlled $C_2$} & \textbf{$\Delta$} \\
\midrule
6 non-reasoning models (headline) & 51.9\,\% & \textbf{80.4\,\%} & $+28.4$~pp \\
8 models (grand aggregate)        & 42.7\,\% & 64.2\,\%          & $+21.6$~pp \\
\bottomrule
\end{tabular}}
\caption{Length-controlled bidirectional NLI $C_2$ at L1 Condition~B. L0 truncated to L1-B's wordpiece-token length; same judge, same denominator as the headline.}
\label{tab:length_controlled}
\end{table}

\paragraph{Per-cell pattern.}
Same direction on every non-reasoning `(model, dataset)' cell. Largest shifts: GPT-4o on GSM8K (18.3\% $\to$ 92.0\%, $+73.8$~pp), Sonnet on GSM8K ($+55.8$~pp), Qwen2.5-VL-7B on GSM8K ($+55.0$~pp).

\paragraph{System-prompt noise floor.}
We do not separately score an L0-A vs.\ L0-B paired-NLI baseline that would isolate the system-prompt register shift between conditions; the headline should be read as a floor above any such (unmeasured) noise.

\FloatBarrier
\section{Comparison with LLMLingua-2}
\label{app:llmlingua}

\begin{figure}[!htbp]
\centering
\includegraphics[width=\columnwidth]{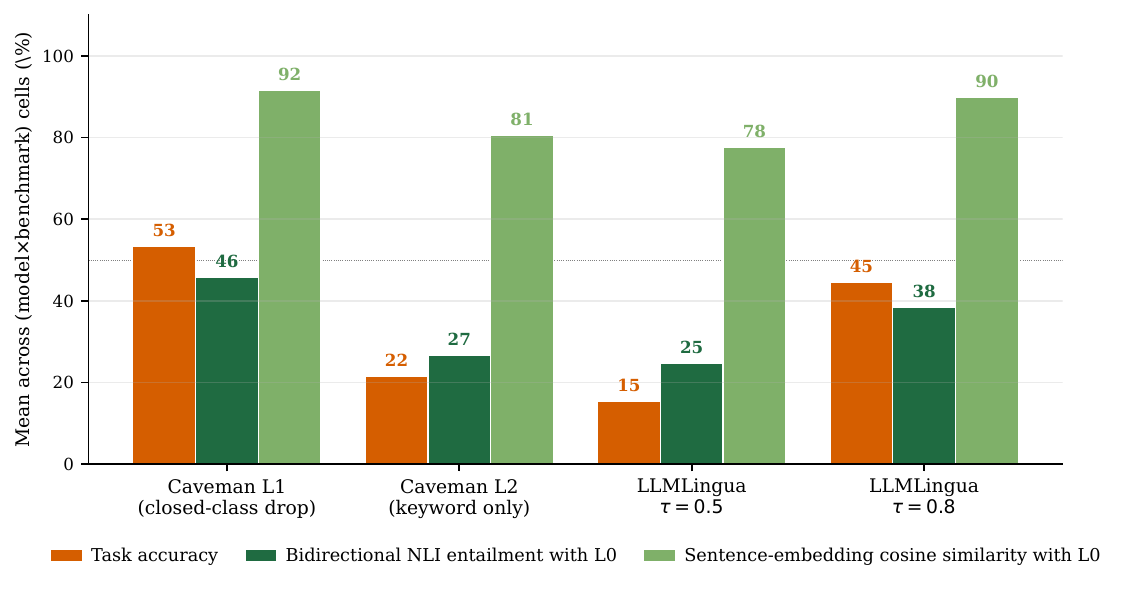}
\caption{Mean accuracy, bidirectional NLI entailment against L0, and dissociation rate ($C_2$) for four input-compression configurations, averaged over the 3 models $\times$ 3 datasets the comparison was run on ($\tau{=}0.8$ is Qwen2.5-VL-7B only; see Table~\ref{tab:llmlingua_summary}). $C_2 > 0$ under every method; the divergence reproduces beyond the POS filter.}
\label{fig:llmlingua_comparison}
\end{figure}

LLMLingua-2 was designed for compressing few-shot demonstrations and long-context inputs, not individual questions in short benchmarks. Our comparison applies it outside its intended regime; we report it because it is the standard learned input-side baseline. We re-ran the input channel under LLMLingua-2 \citep{pan2024llmlingua} on 3 models (GPT-4o, Sonnet~4.6, Qwen2.5-VL-7B) $\times$ 3 datasets (GSM8K, BoolQ, ARC-Easy) at the paper-default $\tau{=}0.5$ rate and at $\tau{=}0.8$ on Qwen2.5-VL-7B (matched to \textsc{Cavewoman}'s telegraphic retention). Table~\ref{tab:llmlingua_summary} and Figure~\ref{fig:llmlingua_comparison} report cross-cell aggregates.

\begin{table}[ht]
\centering
\small
\setlength{\tabcolsep}{6pt}
\resizebox{\columnwidth}{!}{%
\begin{tabular}{lccc}
\toprule
\textbf{Method} & \textbf{Mean acc.} & \textbf{Mean NLI} & \textbf{Mean $C_2$} \\
\midrule
\textsc{Cavewoman} (telegraphic)        & 0.53 & 0.46 & 0.27 \\
\textsc{Cavewoman} (keyword)            & 0.22 & 0.27 & 0.14 \\
LLMLingua-2 ($\tau = 0.5$)     & 0.15 & 0.25 & 0.08 \\
LLMLingua-2 ($\tau = 0.8$)$^{\dagger}$     & 0.45 & 0.38 & 0.22 \\
\bottomrule
\end{tabular}}
\caption{Input compression, 3 models $\times$ 3 datasets at $\tau{=}0.5$; $\tau{=}0.8$ was run on Qwen2.5-VL-7B only as a rate-sensitivity check at a retention rate matched to \textsc{Cavewoman}'s telegraphic level. $C_2 > 0$ on every cell.}
\label{tab:llmlingua_summary}
\end{table}

Three results. (i) The divergence generalizes beyond our POS filter: $C_2 > 0$ on every cell under both methods. (ii) LLMLingua-2's published GSM8K robustness does not transfer to question-text compression: GSM8K accuracy at the default rate is 20--35\% across the three models, well below the $\sim$79\% they report for few-shot demonstration compression. (iii) The collapse is rate-driven, not method-driven: at $\tau{=}0.8$, LLMLingua-2 on Qwen2.5-VL-7B \emph{GSM8K} reaches $0.66$ accuracy and $0.73$ NLI rate, comparable to \textsc{Cavewoman}'s telegraphic level (the lower $\tau{=}0.8$ values in Table~\ref{tab:llmlingua_summary} are the three-dataset Qwen2.5-VL-7B mean).

\paragraph{Compressed prompt at $\tau{=}0.5$.}
Same GSM8K question (Janet's ducks, ground-truth answer $18$) before and after LLMLingua-2 compression. Articles, prepositions, and ``how much'' phrasing are removed; the model often fails. The $\tau{=}0.8$ check preserves 51 of 64 tokens and recovers most of the accuracy gap; per-cell numbers are released in our repository.

\begin{tcolorbox}[
  enhanced, colback=gray!10, colframe=gray!55, colbacktitle=white, titlerule=0pt, boxrule=0.5pt, arc=2pt,
  title={\small\textbf{Original prompt}\quad (64 tokens)},
  coltitle=black, left=5pt, right=5pt, top=3pt, bottom=3pt]
{\small\ttfamily
Janet's ducks lay 16 eggs per day. She eats three for breakfast every morning and bakes muffins for her friends every day with four. She sells the remainder at the farmers' market daily for \$2 per fresh duck egg. How much in dollars does she make every day at the farmers' market?
}
\end{tcolorbox}

\begin{tcolorbox}[
  enhanced, colback=red!4, colframe=red!55!black, colbacktitle=white, titlerule=0pt, boxrule=0.5pt, arc=2pt,
  title={\small\textbf{LLMLingua-2 at $\tau{=}0.5$}\quad (30 tokens; $\approx 47\%$ kept)},
  coltitle=black, left=5pt, right=5pt, top=3pt, bottom=3pt]
{\small\ttfamily
Janet ducks lay 16 eggs. eats three breakfast bakes muffins. sells remainder farmers ' market \$ 2 fresh duck egg. dollars?
}
\end{tcolorbox}

\paragraph{Structural failure mode at the default rate.}
LLMLingua-2 prunes single-letter MCQ labels (``A:'', ``B:'', \ldots) as low-information tokens, collapsing ARC-Easy accuracy to 6--7\% across all three models at $\tau{=}0.5$. \textsc{Cavewoman}'s POS filter retains these tokens by construction.

\FloatBarrier
\section{Reasoning-Token Accounting (DeepSeek-R1 and GPT-5.4)}
\label{app:reasoning_tokens}

DeepSeek-R1-Distill-Qwen-7B emits hidden \texttt{<think>...</think>} traces before its visible answer; HuggingFace \texttt{generate} counts these against \texttt{max\_new\_tokens}, so the per-level decoder budget covers the trace and the visible response together. The trace counts as \texttt{output\_tokens} and enters the NLI judge when it survives. At L4 (20-subword cap) the budget is exhausted before the visible answer starts, which is why DeepSeek-R1 collapses to near-zero accuracy at L4 on every dataset (Table~\ref{tab:acc_summary_B}); at L1--L3, partial traces bias bidirectional NLI against DeepSeek-R1 relative to non-reasoning models.

The pattern also inverts DeepSeek-R1's accuracy-vs-reference-text cells: $C_2$ is small ($\sim$19\%) on the DeepSeek-R1 L1 panel and $C_3$ (agreement despite incorrect) is comparatively large ($\sim$21\%) there. The reported output/input accuracy ratio of $2.4$ is therefore not apples-to-apples with the non-reasoning panel, since the visible budget is effectively shorter.

GPT-5.4 exhibits the same accounting on the API side. Its billed \texttt{output\_tokens} include server-side reasoning tokens that never appear in the returned text: at L0 Condition~B the visible response averages $\sim$56 tokens on BoolQ while the billed output implies $\sim$198, a $3.5\times$ gap on that benchmark ($2.9\times$ averaged across the five benchmarks); GPT-4o, Haiku, and Sonnet bill exactly their visible output. Because the visible-output constraint shortens only the returned text and not the hidden reasoning, output compression leaves GPT-5.4's priced token count essentially unchanged, while the L1 telegraphic system prompt adds $\sim$105 input tokens; its realized cost is therefore flat to slightly higher under output compression (Finding~1). Unlike DeepSeek-R1, these reasoning tokens are not part of the visible text and so do not enter the bidirectional NLI judge, so GPT-5.4's surface-text divergence (Finding~2) is measured on visible text alone and is unaffected.

A third reasoning-protocol model, Kimi-K2.6, was instrumented in the evaluation but excluded from the eight-model panel for a related reason: its constrained-output responses (L1 Condition~B and stricter) returned empty visible text on $99$--$100\%$ of items, with token consumption equal to the per-level \texttt{max\_new\_tokens} cap. The reasoning-block protocol consumes the entire budget under output constraint, the dual of the DeepSeek-R1 case: where DeepSeek-R1's trace partially survives and biases entailment, Kimi's consumes everything and leaves nothing visible. We note it because the failure mode marks a boundary of the output-constraint protocol for reasoning-block models.

\FloatBarrier
\section{Limitations in Detail}
\label{app:limitations_detail}

\paragraph{Length and register confound.}
L1--L3 output-compression generations are forced into a shorter, function-word-light register than L0, and the DeBERTa-NLI training distribution does not cover (telegraphic, prose) pairs. The length-controlled re-scoring of Appendix~\ref{app:noise_floor} bounds this; BART-MNLI (60.3\% inter-judge agreement, 45.7\% both-fail at L1-B) and answer-anchored NLI corroborate.

\paragraph{Judge reliability above L1.}
See Appendix~\ref{app:judge_reliability}. L1 is judge-reliable; L2 is supplementary; L3 is descriptive only. L4 is excluded.

\paragraph{Reference noise and multiple-comparisons correction.}
The divergence is measured against one greedy L0 draw; sampled-decoding stochasticity is not measured. The L0-A vs.\ L0-B noise floor is not separately measured. Across the 320-cell Wilcoxon family at $\alpha=0.05$, 278 of 320 cells are significant uncorrected and 278 remain after Benjamini--Hochberg correction (zero lost). Per-cell bootstrap CIs ($n_{\text{boot}}=1{,}000$) span 1--7~pp.

\FloatBarrier
\section{Extraction-Rate Audit and L4 Length Distribution}
\label{app:extraction_audit}

\subsection{Extraction-Rate Audit}

Accuracy is the fraction of items whose extracted answer matches the ground-truth answer; un-extracted items count as incorrect. When L0 parse rate is materially below 1.0, ``L1 exceeds L0'' partially reflects the extractor working better on shorter outputs.

\paragraph{Headline gap on MMLU-STEM (Condition B).}
Among the models showing apparent L1 gains, L0-B MMLU-STEM parse rate ranges from $0.492$ (Gemma-4-E4B) to $0.854$ (Sonnet) and recovers to $0.807$--$0.969$ at L1-B. Largest parse-rate gaps: Gemma-4-E4B ($+38.7$~pp parse rate), GPT-4o ($+16.8$~pp parse rate), Qwen2.5-VL-7B ($+14.1$~pp parse rate). We do not report L1-vs-L0 accuracy gains where L0 parse $<0.95$.

\paragraph{Other affected cells.}
45 of 80 `(model, dataset, condition)' cells have L0 parse $<0.95$. Sonnet L0-A CSQA is the most extreme ($0.037$) and drives the apparent ``Sonnet L1-A exceeds L0-A by $5$~pp on CSQA'' artifact in Table~\ref{tab:acc_summary_A}. Qwen3.5-9B's below-random L0-B accuracies on ARC-Easy, CommonsenseQA, and MMLU-STEM are not explained by the extraction-rate evidence reported here. The full per-cell audit is released in our repository.

\subsection{L4 Output-Length Distribution}

L4's 15-token target is conveyed through the prompt with a $\texttt{max\_new\_tokens}=20$ ceiling (Table~\ref{tab:l4_length_grand}).

\begin{table}[ht]
\centering
\footnotesize
\setlength{\tabcolsep}{6pt}
\resizebox{\columnwidth}{!}{%
\begin{tabular}{lrr}
\toprule
\textbf{Metric} & \textbf{L4-A (budget 15, soft)} & \textbf{L4-B (budget 20)} \\
\midrule
$n$ items                            & 91{,}720 & 91{,}720 \\
Weighted mean output tokens          & 19.8     & 9.9      \\
Weighted violation rate ($>15$ A, $>20$ B) & 99.0\%   & 0.0\%    \\
\bottomrule
\end{tabular}}
\caption{L4 output-length distribution (40 model$\times$dataset cells per condition). Cond~A overshoots the 15-token target; the same ceiling under Cond~B binds tightly.}
\label{tab:l4_length_grand}
\end{table}

L4-A is retained in per-level tables with the soft-constraint caveat: ``asked to compress to 15 tokens but allowed up to 20'' rather than a strict 15-token budget.

\FloatBarrier
\section{Constraint-Level Specifications}
\label{app:condA}

Verbatim system prompts and POS-tag filter rules for both conditions; the implementing code is released in our repository.

\subsection{Condition A: Input-Compression Filter}

System prompt is fixed across all five levels; only the user message changes via a deterministic spaCy POS-tag filter. Surviving tokens are rejoined with single whitespace; an empty filter output falls back to the original text.

\paragraph{Neutral system prompt (identical at all levels).}

\begin{tcolorbox}[
  colback=gray!8, colframe=gray!45, boxrule=0.5pt, arc=2pt,
  title={\small\textbf{Condition A: System Prompt (L0--L4)}},
  coltitle=black, fonttitle=\bfseries,
  left=4pt, right=4pt, top=2pt, bottom=2pt]
\begin{lstlisting}[style=prompt]
You are a helpful assistant. Answer the following question accurately and completely.
\end{lstlisting}
\end{tcolorbox}

\paragraph{POS-tag filter rules.}

Table~\ref{tab:posrules} gives the per-level filter rule applied to the user message. The L4 row truncates the \emph{token stream} (not the character sequence), preserving the monotone ladder $\text{L0}\supseteq\text{L1}\supseteq\text{L2}\supseteq\text{L3}\supseteq\text{L4}$.

\begin{table*}[!htbp]
\centering
\small
\begin{tabular}{ll}
\toprule
Level & Rule \\
\midrule
L0 & unchanged (original text) \\
L1 & drop \texttt{DT}, \texttt{IN}, \texttt{CC}, \texttt{RP}, \texttt{TO}, \texttt{MD} \\
L2 & keep \texttt{NN}, \texttt{NNS}, \texttt{NNP}, \texttt{NNPS},
     \texttt{VB}, \texttt{VBZ}, \texttt{VBD}, \texttt{VBN}, \texttt{VBG}, \texttt{CD} \\
L3 & keep \texttt{NN}, \texttt{NNS}, \texttt{NNP}, \texttt{NNPS}, \texttt{CD} \\
L4 & apply L3 filter, then keep first 15 whitespace tokens \\
\bottomrule
\end{tabular}
\caption{Condition~A per-level POS-tag filter rules.}
\label{tab:posrules}
\end{table*}

\begin{figure*}[!htbp]
\centering
\includegraphics[width=\textwidth]{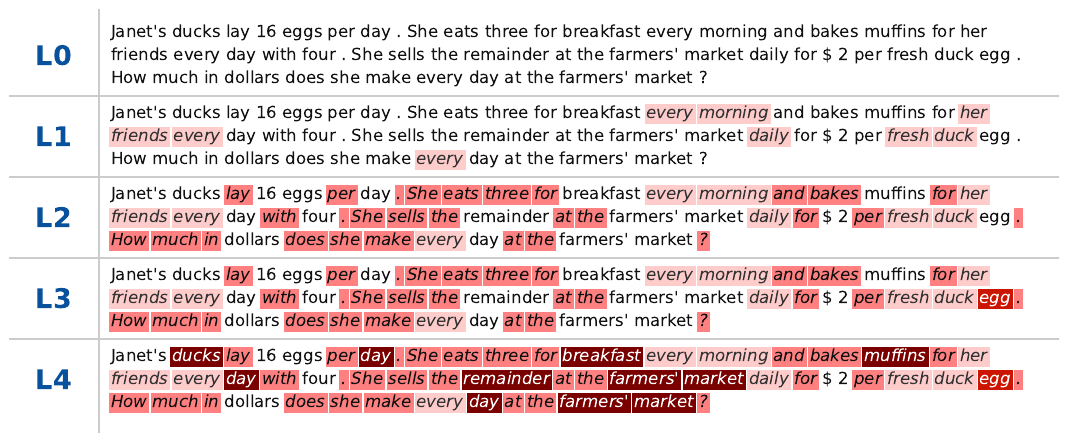}
\caption{Worked example of the input-compression filter applied to a single question at each of the five reduction levels. L0 is the unmodified prompt; L1 removes closed-class function words; L2 retains nouns, verbs, and cardinal numerals; L3 strips the verbs to leave a nominal skeleton; L4 truncates the L3 form to its first fifteen tokens. The same filter family defines both channels.}
\label{fig:input_compression}
\end{figure*}

\subsection{Condition B: Per-Level System Prompts}
\label{app:condB}

Each prompt has a named constraint type, a rule list, a task-neutral example, and the answer-format convention. Decoder budgets: $\{L_0\!:\!400,\ L_1\!:\!300,\ L_2\!:\!200,\ L_3\!:\!150,\ L_4\!:\!20\}$ \texttt{max\_new\_tokens}.

\begin{tcolorbox}[
  colback=gray!8, colframe=gray!45, boxrule=0.5pt, arc=2pt,
  title={\small\textbf{Condition B: L0 (Unconstrained)}},
  coltitle=black, fonttitle=\bfseries,
  left=4pt, right=4pt, top=2pt, bottom=2pt]
\begin{lstlisting}[style=prompt]
Answer the following question accurately.
Reason step by step in full, grammatical English sentences. Conclude with the final answer on its own line in the form 'Answer: <answer>'.

The final-line answer matches what the question asks for: a number for numeric questions, 'yes' or 'no' for yes/no questions, or a single letter (A, B, C, ...) for multiple-choice questions.
\end{lstlisting}
\end{tcolorbox}

\begin{tcolorbox}[
  colback=gray!8, colframe=gray!45, boxrule=0.5pt, arc=2pt,
  title={\small\textbf{Condition B: L1 (Telegraphic)}},
  coltitle=black, fonttitle=\bfseries,
  left=4pt, right=4pt, top=2pt, bottom=2pt]
\begin{lstlisting}[style=prompt]
Answer the question under a TELEGRAPHIC constraint.

Rules:
  - DO NOT use any function words. No articles (the, a, an). No conjunctions (and, but, or, so). No prepositions (of, in, to, for, at, with, from, by, on, per).
  - DO use nouns, main verbs, numbers, and standard symbols (+, -, *, /, =).
  - Show each reasoning step.
  - End with a line: 'Answer: <answer>'.

Example (task-neutral):
  Premise mentions item X. Property Y holds X. Match: yes.
  Answer: <answer>

The final-line answer matches what the question asks for: a number for numeric questions, 'yes' or 'no' for yes/no questions, or a single letter (A, B, C, ...) for multiple-choice questions.
\end{lstlisting}
\end{tcolorbox}

\begin{tcolorbox}[
  colback=gray!8, colframe=gray!45, boxrule=0.5pt, arc=2pt,
  title={\small\textbf{Condition B: L2 (Keyword-Only)}},
  coltitle=black, fonttitle=\bfseries,
  left=4pt, right=4pt, top=2pt, bottom=2pt]
\begin{lstlisting}[style=prompt]
Answer the question under a KEYWORD-ONLY constraint.

Rules:
  - Use ONLY nouns and main verbs. No grammar, no full sentences.
  - Output as fragments, short labels, or list items.
  - Numbers and standard symbols (+, -, *, /, =) are allowed.
  - Each reasoning step appears as a fragment.
  - End with a line: 'Answer: <answer>'.

Example:
  Item: X
  Property Y: holds
  Match: yes
  Answer: <answer>

The final-line answer matches what the question asks for: a number for numeric questions, 'yes' or 'no' for yes/no questions, or a single letter (A, B, C, ...) for multiple-choice questions.
\end{lstlisting}
\end{tcolorbox}

\begin{tcolorbox}[
  colback=gray!8, colframe=gray!45, boxrule=0.5pt, arc=2pt,
  title={\small\textbf{Condition B: L3 (Noun-Phrase Skeleton)}},
  coltitle=black, fonttitle=\bfseries,
  left=4pt, right=4pt, top=2pt, bottom=2pt]
\begin{lstlisting}[style=prompt]
Answer the question under a NOUN-PHRASE SKELETON constraint.

Rules:
  - NO verbs of any kind. None.
  - Use only nominal fragments: nouns, noun compounds, numbers, and standard symbols (+, -, *, /, =).
  - Each step is a noun phrase labeling a quantity, claim, or property.
  - End with a line: 'Answer: <answer>'.

Example:
  Item: X
  Property in question: Y
  Match status: positive
  Answer: <answer>

The final-line answer matches what the question asks for: a number for numeric questions, 'yes' or 'no' for yes/no questions, or a single letter (A, B, C, ...) for multiple-choice questions.
\end{lstlisting}
\end{tcolorbox}

\begin{tcolorbox}[
  colback=gray!8, colframe=gray!45, boxrule=0.5pt, arc=2pt,
  title={\small\textbf{Condition B: L4 (Hard 15-Token Budget)}},
  coltitle=black, fonttitle=\bfseries,
  left=4pt, right=4pt, top=2pt, bottom=2pt]
\begin{lstlisting}[style=prompt]
Answer the question under a HARD TOKEN BUDGET.

Rules:
  - Your ENTIRE response must be 15 tokens or fewer.
  - The response MUST include the final answer.
  - Prefer the raw answer over prose.

Example: 'Answer: <answer>'

The final-line answer matches what the question asks for: a number for numeric questions, 'yes' or 'no' for yes/no questions, or a single letter (A, B, C, ...) for multiple-choice questions.
\end{lstlisting}
\end{tcolorbox}

\FloatBarrier
\section{Qualitative Examples}
\label{app:examples}

One BoolQ item (Qwen2.5-VL-7B, Condition~B) traced across all five levels (L4 is shown for illustration only; it is excluded from the aggregate semantic scoring of \S\ref{sec:metrics}). Box color: \colorbox{green!10}{\strut green = $C_1$ (correct, entailment PASS)}, \colorbox{orange!15}{\strut orange = $C_2$ (correct, entailment FAIL)}, gray = L0. The correct answer survives compression after the reasoning chain has collapsed (the $C_2$ pattern of \S\ref{sec:finding-dissociation}). A GSM8K arithmetic trace is in the released artifact bundle.

\subsection*{BoolQ: Passage Yes/No (Qwen2.5-VL-7B, Condition~B)}

\textbf{Question:} is harry potter and the escape from gringotts a roller coaster ride\\
\textbf{Ground truth answer:} yes\\
\textit{(BoolQ provides a supporting passage with each question; passage omitted here for space.)}

\smallskip

\begin{tcolorbox}[
  enhanced, colback=gray!12, colframe=gray!55, colbacktitle=white, titlerule=0pt, boxrule=0.5pt, arc=2pt,
  title={\small\textbf{L0 (Unconstrained baseline)} \quad Pred: yes \quad GT: yes \quad \textcolor{green!50!black}{Correct}},
  coltitle=black, left=5pt, right=5pt, top=3pt, bottom=3pt]
{\small\ttfamily
To determine if this attraction is a roller coaster ride, I analyzed the passage. The passage states it is an ``indoor steel roller coaster'' at Universal Studios Florida, which directly confirms it is a roller coaster. Based on this evidence\ldots

\textbf{Answer: yes}}
\end{tcolorbox}

\begin{tcolorbox}[
  enhanced, colback=orange!9, colframe=orange!65!black, colbacktitle=white, titlerule=0pt, boxrule=0.5pt, arc=2pt,
  title={\small\textbf{L1 (Telegraphic)} \quad Pred: yes \quad GT: yes \quad \textcolor{green!50!black}{Correct} \quad NLI: \textcolor{red!60!black}{FAIL} \quad [$C_2$]},
  coltitle=black, left=5pt, right=5pt, top=3pt, bottom=3pt]
{\small\ttfamily
Yes.

\textbf{Answer: yes}}
\end{tcolorbox}

\begin{tcolorbox}[
  enhanced, colback=orange!9, colframe=orange!65!black, colbacktitle=white, titlerule=0pt, boxrule=0.5pt, arc=2pt,
  title={\small\textbf{L2 (Keyword-Only)} \quad Pred: yes \quad GT: yes \quad \textcolor{green!50!black}{Correct} \quad NLI: \textcolor{red!60!black}{FAIL} \quad [$C_2$]},
  coltitle=black, left=5pt, right=5pt, top=3pt, bottom=3pt]
{\small\ttfamily Yes}
\end{tcolorbox}

\begin{tcolorbox}[
  enhanced, colback=green!5, colframe=green!55!black, colbacktitle=white, titlerule=0pt, boxrule=0.5pt, arc=2pt,
  title={\small\textbf{L3 (Noun-Phrase Skeleton)} \quad Pred: yes \quad GT: yes \quad \textcolor{green!50!black}{Correct} \quad NLI: \textcolor{green!50!black}{PASS} \quad [$C_1$]},
  coltitle=black, left=5pt, right=5pt, top=3pt, bottom=3pt]
{\small\ttfamily
Item: Harry Potter and the Escape from Gringotts\\
Property in question: Roller coaster ride\\
Match status: positive

\textbf{Answer: yes}}
\end{tcolorbox}

\begin{tcolorbox}[
  enhanced, colback=orange!9, colframe=orange!65!black, colbacktitle=white, titlerule=0pt, boxrule=0.5pt, arc=2pt,
  title={\small\textbf{L4 (Hard 15-Token Budget)} \quad Pred: yes \quad GT: yes \quad \textcolor{green!50!black}{Correct} \quad NLI: \textcolor{red!60!black}{FAIL} \quad [$C_2$]},
  coltitle=black, left=5pt, right=5pt, top=3pt, bottom=3pt]
{\small\ttfamily \textbf{Answer: Yes}}
\end{tcolorbox}

\textit{Takeaway.} The correct answer survives at every level, but the reasoning chain collapses at L1, L2, and L4 (NLI FAIL; $C_2$ cell). The L0 output makes several inferential steps citing the passage; ``Yes.'' and ``Yes'' make no such claims, so the L0 output does not entail them. L3 unexpectedly recovers bidirectional entailment: the noun-phrase skeleton labels the entity and property correctly, replicating enough of the L0 propositional structure for entailment to pass. This item exemplifies why accuracy alone cannot distinguish $C_1$ from $C_2$.

\FloatBarrier
\section{Released Artefacts and Reproducibility}
\label{app:reproducibility}

Each (model, dataset, condition, level) configuration releases three artifacts in our repository: a base inference record (token counts, realized cost, extracted answer, ground-truth answer), a paired entailment record (bidirectional NLI scores at L1--L4), and a paired embedding record (sentence-embedding cosine at L0--L4). Each configuration is accompanied by a run-configuration manifest recording the git SHA, conda environment, package versions, and GPU. Aggregate per-cell summaries and a verification script that reproduces the paper's reported numbers from those summaries are released alongside.

\end{document}